\documentclass{article} 
\usepackage{style/iclr2026_conference,times}

\usepackage{amsmath,amsfonts,bm}

\def\eqref#1{equation~\ref{#1}}

\def\1{\bm{1}}

\DeclareMathAlphabet{\mathsfit}{\encodingdefault}{\sfdefault}{m}{sl}
\SetMathAlphabet{\mathsfit}{bold}{\encodingdefault}{\sfdefault}{bx}{n}

\usepackage[table]{xcolor}
\usepackage{makecell} 
\usepackage{fancyvrb} 
\usepackage{booktabs} 
\usepackage{graphicx}
\usepackage{booktabs}
\usepackage{wrapfig}
\usepackage{array}
\usepackage{tikz}
\usepackage{pgfplots}
\usepackage{colortbl}
\usepackage[accsupp]{axessibility}  %
\usepackage{orcidlink}
\usepackage{tabularx}
\usepackage{multirow}
\usepackage{tcolorbox}
\usepackage{tikz}
\usepackage{pifont}
\usepackage{microtype}
\usepackage{fontawesome}
\usepackage[switch]{lineno} 
\usepackage[toc,page]{appendix}
\usepackage{bm}
\usepackage{gradient-text}
\usepackage{textcomp}
\usepackage{gensymb}
\usepackage{lipsum}
\usepackage{xspace}

\usepackage{caption}
\usepackage{times}  
\usepackage{url}
\usepackage{makecell} %
\usepackage{wrapfig}

\usepackage{listings}
\newcommand{\qheading}[1]{\noindent\mbox{\textbf{#1}}}

\definecolor{bestgreen}{RGB}{153,200,76}
\definecolor{worstred}{RGB}{192,0,0}

\definecolor{cbad}{HTML}{FFD0D0} 
\definecolor{cmedium}{HTML}{FFF0D0}
\definecolor{cgood}{HTML}{90C060}

\usepackage{pgfplots}
\pgfplotsset{compat=1.18}

\newlength\savewidth

\definecolor{DeltaColor}{rgb}{0.039,0.73,0.71}
\definecolor{SigmaColor}{rgb}{0.98,0.45,0.0}
\definecolor{AlphaColor}{rgb}{0,0,0.8}
\definecolor{BetaColor}{rgb}{0.8,0,0.8}
\definecolor{GammaColor}{rgb}{0.514,0.34,0.224}
\definecolor{EpsilonColor}{rgb}{0.353,0.725,0.906}
\definecolor{PurpleColor}{HTML}{9839ff}
\definecolor{RedColor}{rgb}{0.949,0.275, 0.224}
\definecolor{citecolor}{HTML}{0071bc}

\usepackage[utf8]{inputenc}

\newcommand{\longtitle}{Character Mixing for Video Generation}
\newcommand{\ourtitle}{\longtitle}

\DefineVerbatimEnvironment{PromptBlock}{Verbatim}
{fontsize=\small, 
 baselinestretch=1.0, 
 frame=single, 
 rulecolor=\color{gray}, 
 breaklines=true,       
 breakanywhere=true,    
}

\definecolor{headergray}{RGB}{245, 245, 245}
\definecolor{baselineblue}{RGB}{235, 242, 250}
\definecolor{highlightgold}{RGB}{255, 250, 220}

\newcommand{\page}{\href{https://tingtingliao.github.io/mimix}{\textcolor{magenta}{\xspace{https://tingtingliao.github.io/mimix}}\xspace}}

\newcommand{\mrbean}{\textit{Mr. Bean}}
\newcommand{\tomandjerry}{\textit{Tom and Jerry}}
\newcommand{\threebears}{\textit{We Bare Bears}}
\newcommand{\youngsheldon}{\textit{Young Sheldon}}

\definecolor{PurpleColor}{HTML}{8B008B}
\definecolor{OrangeColor}{rgb}{0.914,0.541,0.0.141}
\definecolor{GreenColor}{rgb}{0.137,0.573,0.565}

\let\cite\citep
\usepackage{hyperref}
\hypersetup{
    colorlinks=true,
    linkcolor=red,
    citecolor=blue,
    urlcolor=teal
}
\title{\ourtitle}

\author{
Tingting Liao \quad Chongjian Ge \quad Guangyi Liu \quad Hao Li \quad Yi Zhou \\
Mohamed bin Zayed University of Artificial Intelligence 
}

\iclrfinalcopy 
\begin{document}

\maketitle
\begin{figure}[h]
 
    \begin{minipage}{\linewidth}
        \scriptsize
        \raggedright
        \textbf{Prompts-1:} \textit{\textbf{Ice Bear} calmly paints a picture of \textbf{Tom}, while \textbf{Tom} keeps trying to pose but falls into the paint buckets.} 
        
        \textbf{Prompts-2:} \textit{\textbf{Mr. Bean} blows up a balloon. \textbf{Jerry} hides inside. When the balloon pops \textbf{Jerry} lands on \textbf{Mr. Bean’s} head}. 
        
        \textbf{Prompts-3:} \textit{\textbf{Young Sheldon} judges a spelling bee, \textbf{Panda} spells words wrong on purpose, while \textbf{Jerry} sneaks in funny answers.} 
        
    \end{minipage}
    \includegraphics[width=\textwidth]{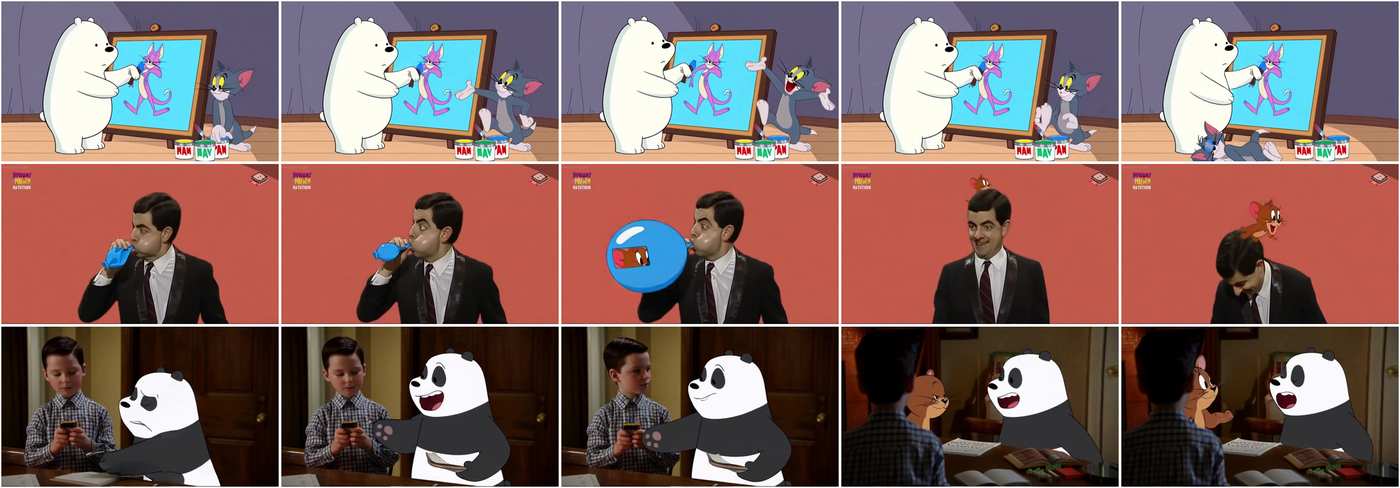}

\caption{
    \textbf{Multi-character Mixing.} Our method preserves character identity, behavior and original style while generating plausible interactions between characters that have never coexisted—from cartoons (\threebears, \tomandjerry) to realistic humans (\mrbean, \youngsheldon).
}

\label{fig:teaser}
\end{figure}

\begin{abstract}

    Imagine Mr. Bean stepping into \textit{Tom and Jerry}---can we generate videos where characters interact naturally across different worlds? We study inter-character interaction in text-to-video generation, where the key challenge is to preserve each character's identity and behaviors while enabling coherent cross-context interaction. This is difficult because characters may never have coexisted and because mixing styles often causes \textbf{\textit{style delusion}}, where realistic characters appear cartoonish or vice versa. We introduce a framework that tackles these issues with Cross-Character Embedding (CCE), which learns identity and behavioral logic across multimodal sources, and Cross-Character Augmentation (CCA), which enriches training with synthetic co-existence and mixed-style data. Together, these techniques allow natural interactions between previously uncoexistent characters without losing stylistic fidelity. Experiments on a curated benchmark of cartoons and live-action series with 10 characters show clear improvements in identity preservation, interaction quality, and robustness to style delusion, enabling new forms of generative storytelling. 
    Additional results and videos are available on our project page: \page.

\end{abstract}    
\section{Introduction}
In an era where films and iconic characters are just a click away, a natural question arises: what if we could unite these beloved characters together, merging their roles and interactions into a single story?

Since the release of Sora~\cite{sora} by OpenAI, fundamental text-to-video (T2V) generation models~\cite{cogvideox,sora,wan2025video,genmo2024mochi,google2024veo,kong2024hunyuanvideo,kling,wan2025video} have achieved substantial progress in general video synthesis. For producing videos centered on specific or customized characters, a common approach is to leverage a reference image as input. Such personalized video generation methods~\cite{he2024idanimator,liang2025movieweaver,videobooth,dreamvideo,fei2025skyreelsa2,chen2025videoalchemist} allow users to create customized content with their own images. 
However, reference images alone provide visual similarity but fail to capture motion characteristics such as the character's distinctive behaviors and interactions with the environment, and others.
Thus, while preserving the identity, these approaches often fail to faithfully capture the character's complex behavior.
To this end, we want to develop a method that could allow the model to learn from all the footages and scripts of the characters of interest that can be collected, so as to learn not only their appearance but also motion idiosyncrasies and personality, and to generate vivid videos that maximally match their behavior traits and motion patterns.

To faithfully generate videos for a single character, one can fine-tune a text-to-video model on footage of that character. However, moving beyond single-character settings brings two major challenges. 

The first is the \textbf{non-coexistence challenge}: characters from different shows never co-occur in any training video, leaving no paired data to model their joint interactions. To address this, we explicitly encode each character’s identity and behavior into text by annotating their names and actions in the captions. This disentangles character-specific behavior embeddings from the underlying training videos, enabling us to fine-tune one foundation model on each character’s individual footages while still allowing them to co-exist and interact naturally at inference time. 

\begin{wrapfigure}{r}{0.5\textwidth}
    \centering
    \vspace{-10pt}  
      
    \includegraphics[width=0.5\textwidth]{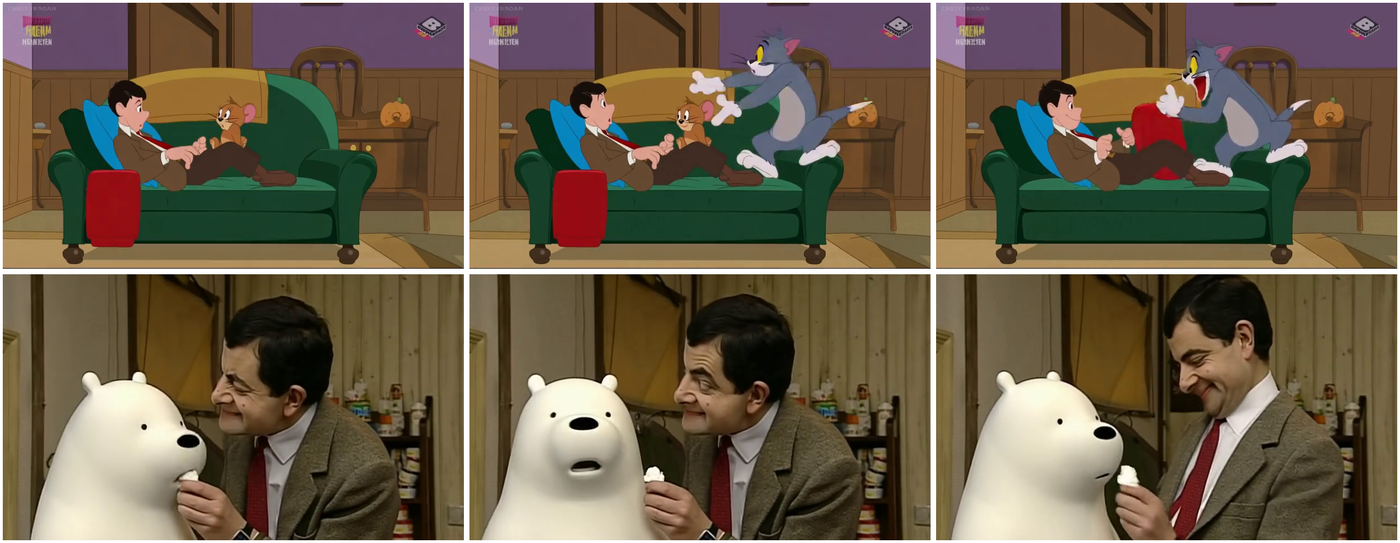}
    
    \caption{\textbf{Style delusion examples.} When mixing different style characters, their styles may shift undesirably. For instance, Mr. Bean looks cartoonish (top row), while Ice Bear appears realistic (bottom row).}
    \label{fig:style_delusion}
    \vspace{-10pt}  
\end{wrapfigure}

The second is the \textbf{style delusion challenge}:
characters often originate from domains with drastically different visual styles, such as live-action sitcoms and cartoons, which never naturally co-exist in the same videos.
Directly training on mixed styled data leads to unstable character styles, as shown in Figure \ref{fig:style_delusion}, where the model tries to render all characters in a uniform style matching the training data.

We tackle this by introducing a style-aware data augmentation strategy that composites characters from different domains into the same video while preserving their native appearances. We find that even a small proportion of such augmented data substantially improves style preservation in cross-domain character mixing. For background ambiguities, we introduce an extra prompt for background style.

To validate our approach, we curate an 81-hour (52,000 clips) dataset featuring two cartoons (\tomandjerry, \threebears) and two realistic shows (\youngsheldon, \mrbean). Each clip is annotated with explicit character names and style information, supporting fine-grained control during training and inference. We further establish the first benchmark for multi-character video generation, evaluating identity preservation, motion fidelity, interaction realism, and style consistency.

Our contributions are summarized as follows:
\begin{itemize}
    \item We proposed the first video generation framework for multi-character mixing that addresses both the non-coexistence challenge and the style dilution challenge.
    \item We curated an 81-hour  (52,000 clips)  dataset with character- and style-annotated captions, enabling controllable multi-character video synthesis across domains.
    \item We conducted extensive experiments and introduced a benchmark, showing that our method significantly improves identity preservation, motion consistency, and interaction quality compared to prior art.
\end{itemize}

\section{Related Works} 

\qheading{Video Generation} The advent of diffusion models has transformed video generation, advancing from early text-to-video systems such as ImagenVideo~\cite{ho2022imagenvideo}, Make-A-Video~\cite{singer2022makeavideo}, and VideoLDM~\cite{blattmann2023videoldm}, to large-scale architectures like Sora \cite{openai2024sora}, Goku~\cite{chen2025goku}, Wan2.1~\cite{wan2025video} and HunyuanVideo~\cite{kong2024hunyuanvideo}, which achieve state-of-the-art general video synthesis. However, these models remain limited in generating content with specific identities or custom subjects, motivating the emerging direction of personalized video generation, where reference visual signals guide the synthesis of videos with consistent appearance and motion dynamics.

\qheading{Single-Concept Personalization} Early personalized video generation methods, such as DreamVideo~\cite{dreamvideo}, Magic-Me~\cite{magicme}, and PersonalVideo~\cite{li2024personalvideo}, customized videos with per-subject tuning, achieving identity preservation but requiring costly optimization. More recent zero-shot approaches like ID-Animator~\cite{he2024idanimator} leverage facial adapters to enable identity-consistent generation from a single reference image without fine-tuning. Despite these advances, existing methods largely emphasize visual similarity while overlooking motion-related aspects such as unique behaviors and environment interactions. 

\qheading{Multi-Concept Customization} Compared to sigle-concept personalization, multi-concept customization is more challenging due to identity blending, where multiple characters risk being fused into a composite scene. Video Alchemist~\cite{chen2025videoalchemist} addresses this through cross-attention–based fusion of text and image representations for open-set subject and background control, while Movie Weaver~\cite{liang2025movieweaver} employs tuning-free anchored prompts to preserve distinct identities. Other approaches, such as CustomVideo~\cite{wang2024customvideo} and Custom Diffusion~\cite{kumari2023customdiffusion}, explore parameter-efficient fine-tuning and joint optimization for multi-subject composition. However, existing image-guided methods cannot leverage video and textual data during training, limiting their ability to model realistic interactions and dynamics across characters.

\section{Methods}

The goal of our method is to learn the essence of characters from large collections of video data and enable them to interact seamlessly in new, mixed contexts. Given the abundance of video series—spanning cartoons and live-action shows, our approach seeks to (1) capture characters' unique identity and behavioral traits, and (2) enable flexible mixing of characters across styles and universes.

Our curated dataset (Section~\ref{sec:dataset_curation}) consists of TV shows and animations. We leverage not only video clips but also audio and scripts, which provide crucial cues about each character’s personality and behavioral logic. Each domain features one or multiple central characters, sometimes appearing in ensembles (e.g., Tom and Jerry), other times in isolation (e.g., Mr. Bean).

In this section, we develop a novel training scheme that introduces \textbf{Cross-Character Embedding (CCE)} and \textbf{Cross-Character Augmentation (CCA)} to achieve robust identity modeling, behavior preservation, and style-controllable mixing.

\subsection{Cross-Character Embedding (CCE)}

Faithfully reproducing a character requires learning from dynamic data that captures not only appearance but also behavior, idiosyncratic motion patterns, and contextual habits. However, images are insufficient because they omit the motion and interaction cues that define a character's identity.

We therefore design a framework to learn the character concept embeddings across different domains. We need to tackle character disentanglement in multi-character shows (e.g., \tomandjerry, \threebears), where multiple identities must be separated within the same clip. We also face the challenge of \textbf{non-coexistence} of characters from different universes, who never appear together in the training data but must interact coherently at inference.  

\begin{figure}[t] 
    \centering 
    \includegraphics[width=\linewidth]{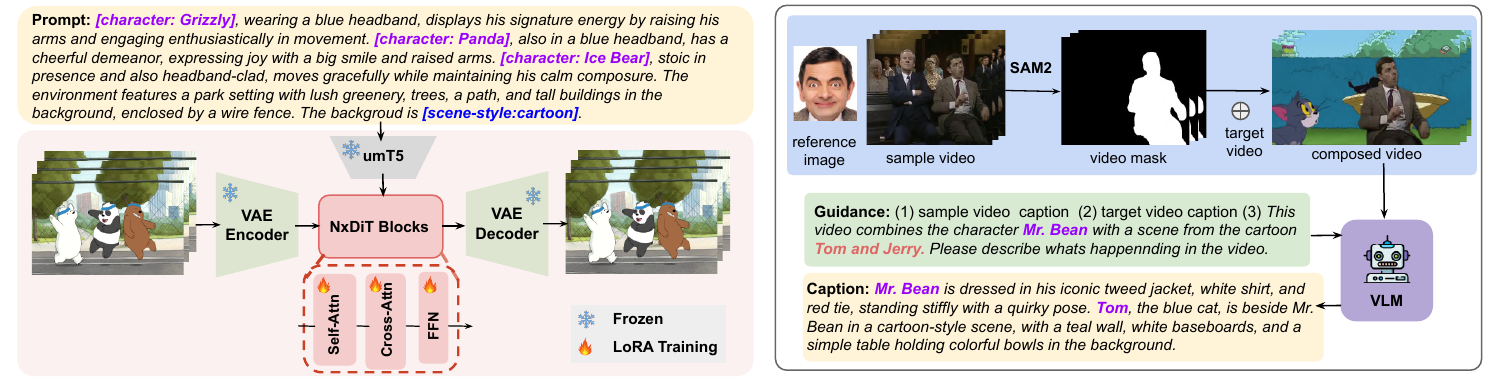} 
    \caption{\textbf{Model finetuning architecture} (left) and \textbf{data augmentation pipeline} (right).}

    \label{fig:method}
\end{figure}

\qheading{Character--Action Prompting.}
Our key insight is to design a character--action captioning format that explicitly grounds each character’s identity while separating it from scene context. Unlike standard captions that describe only visual events, our captions follow the format:  

\begin{center}
\textit{[Character: \textless name\textgreater], \textless action\textgreater. 
[Character: \textless name\textgreater], \textless action\textgreater. }
\end{center}

This design ensures that embeddings encode characters as independent entities with disentangled actions and identities. During inference, the same prompting scheme enables coherent composition of characters who never coexisted in training—for example, Mr. Bean interacting with Tom and Jerry. 
Although Mr. Bean and Jerry never meet each other in the dataset, the model has observed how each interacts with others in their respective domains, allowing it to generalize to cross-universe interactions. A more detailed illustration is provided in Figure~\ref{fig:method}.

\qheading{Prompt Generation.}
We employ GPT-4o~\cite{openai2024gpt4o} to automatically generate captions. For each short clip, we provide: 1) 10 sampled frames as visual context, 2) dialogue transcripts from the audio, and 3) source metadata (cartoon vs. TV series).

Scripts and plot summaries are also supplied to resolve ambiguity when a short clip lacks context. This setup enables GPT-4o to reliably identify character names and actions while minimizing hallucinations. The resulting process yields ~52,000 video–caption pairs. Each character mention is annotated with \texttt{[character:name]} tags, which act as identity anchors and support controllable character-level generation. More details are provided in the supplemental.

\qheading{Model Adaptation.}
We fine-tune Wan2.1-T2V-14B~\cite{wan2025video} with Low-Rank Adaptation (LoRA)~\cite{hu2021lora}. Our approach is model-agnostic and applicable to any text-to-video backbone. The curated text–video pairs capture each character’s actions, emotions, and behaviors, enabling the learned embeddings to serve as flexible building blocks for multi-character generation.

\subsection{Cross-Character Augmentation (CCA)}
While CCE ensures characters act authentically, training across mixed domains (cartoon and live-action) introduces a \textbf{style delusion problem}: characters may drift into unintended visual styles (e.g., \mrbean\xspace rendered as a cartoon, or Ice Bear appearing overly realistic), as shown in Figure~\ref{fig:style_delusion}. 

To preserve original styles while allowing cross-style interactions, we introduce \textbf{Cross-Character Augmentation (CCA)}. This tackles a second non-coexistence challenge: in the training data, cartoon and real characters never appear together, and neither do their backgrounds.  

\qheading{Synthetic Cross-Domain Compositing.}
Our intuition is that even imperfect synthetic co-occurrences can effectively guide the model toward more consistent, style-preserving generation. We therefore create augmented training clips by segmenting characters from source videos and pasting them into backgrounds from the opposite style domain. For example, Mr.Bean (live-action) may be placed in a cartoon \tomandjerry scene as shown in the right part of Figure~\ref{fig:method}.  

Characters are segmented using SAM2~\cite{ravi2024sam2}, which handles both live-action and animation content. To ensure the quality and relevance of the synthetic compositions, we apply different filtering strategies. For live-action characters such as \mrbean and \youngsheldon, we perform reference-image matching to retain clips with accurate identity alignment. For cartoon characters, we leverage Gemini~\cite{gemini2024} for automated detection and filtering of characters. 

The composited clips are then captioned with GPT-4o, which receives both the background source and the inserted character identities. Each caption is further enriched with explicit style tags (\texttt{[scene-style:cartoon]} or \texttt{[scene-style:realistic]}), providing the model with clear supervision for style control. The complete caption format becomes:

\begin{center}
\textit{[Character: \textless name\textgreater], \textless action\textgreater. 
[Character: \textless name\textgreater], \textless action\textgreater. 
\textless scene-style\textgreater}
\end{center}

\qheading{Empirical Findings.}
We observe that a \textbf{small proportion} of such augmented clips suffices to unlock robust cross-style composition. Excessive synthetic data, however, degrades realism and harms overall video quality. A detailed analysis is provided in the experiments section.  

\subsection{Training and Data}
During fine-tuning, backbone parameters are frozen and only LoRA layers are updated, ensuring efficiency and reducing overfitting. We adopt rank-32 LoRA layers and train for 5 epochs with the Adam optimizer (learning rate 1e-4, batch size 64). Gradient clipping is applied for stability, and mixed-precision (FP16) training is used for efficiency.

\qheading{Scenes and segments.}
\label{sec:dataset_curation}
We curate a dataset comprising two cartoons and two live-action shows: approximately 9 hours of \tomandjerry, 18 hours of \threebears, 8 hours of \mrbean, and 46 hours of \youngsheldon. We standardize all videos by cropping out bottom text overlays (e.g., subtitles, credits) to prevent spurious language cues. Videos are segmented scene-by-scene into 5-second clips with the length of 81 frames, at 16 fps. 

For each domain, we define the set of key characters: \mrbean\, (\texttt{Mr Bean}), \tomandjerry\, (\texttt{Tom}, \texttt{Jerry}, \texttt{Spike}), \threebears\, (\texttt{Ice Bear}, \texttt{Grizzly}, \texttt{Panda}), and \youngsheldon\, (\texttt{Sheldon}, \texttt{Missy}, \texttt{Mary Cooper}, \texttt{George Cooper}).

\begin{figure}[t] 
    \centering

    \begin{minipage}{\linewidth}
        \scriptsize
        \raggedright
        \textbf{Prompt:} \textit{\textbf{Tom}, and \textbf{Panda} go fishing on a rowboat. \textbf{Tom} keeps falling into the water while chasing his bait, \textbf{Panda} takes selfies with each fish, and \textbf{Tom} somehow catches a boot, a sandwich, and a lawn chair—but no fish. }
    \end{minipage}
    \includegraphics[width=\linewidth]{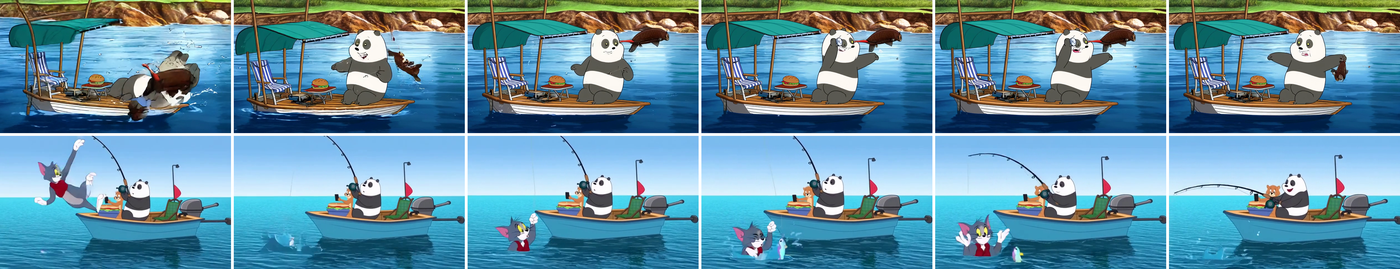}
    
    \begin{minipage}{\linewidth}
        \scriptsize
        \raggedright
        \textbf{Prompt:} \textit{\textbf{Tom} and \textbf{Ice Bear} get jobs at a bakery. \textbf{Tom} keeps chasing \textbf{Jerry} through cake trays, and \textbf{Ice Bear} calmly decorates a three-tier wedding cake. The bride ends up choosing \textbf{Ice Bear}’s version over the original. } 
    \end{minipage}
    \includegraphics[width=\linewidth]{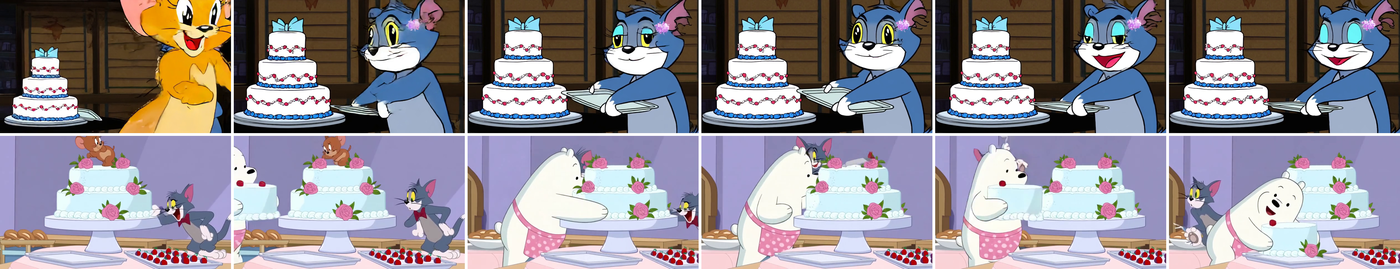}
      
    \begin{minipage}{\linewidth}
        \scriptsize
        \raggedright
        \textbf{Prompt:} \textit{\textbf{Mary Cooper} and \textbf{Panda} host a tea party. \textbf{Mary} brings her finest china and proper etiquette. \textbf{Panda} makes cute cupcakes with bear faces.}  
    \end{minipage}
    \includegraphics[width=\linewidth]{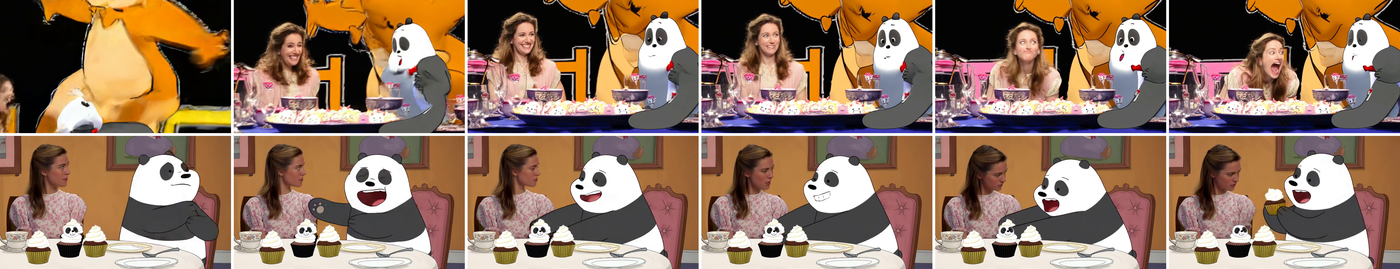}
    
    \begin{minipage}{\linewidth}
        \scriptsize
        \raggedright
        \textbf{Prompt:} \textit{\textbf{Tom} and \textbf{Sheldon} visit an aquarium. \textbf{Tom} gets mesmerized by the fish and tries to pounce on the glass. \textbf{Sheldon} gives lectures to confused toddlers about bioluminescent jellyfish.}  
    \end{minipage}
    \includegraphics[width=\linewidth]{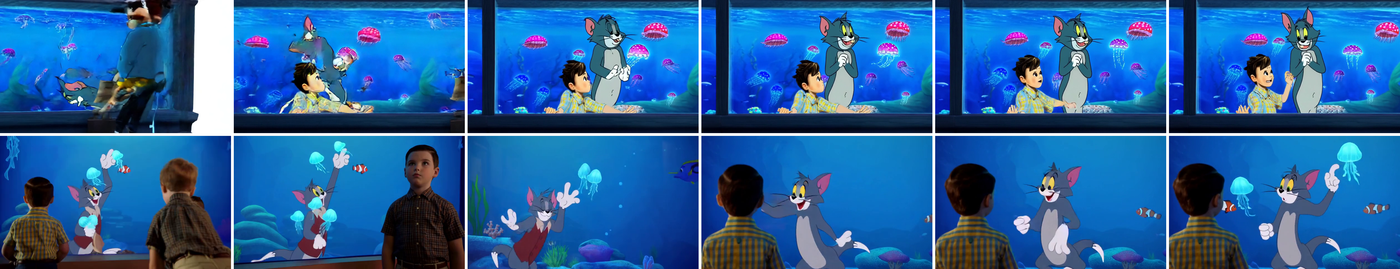}
     
    \caption{\textbf{Comparison on multi-subject interaction.} Results from SkyReel-A2~\cite{fei2025skyreelsa2} (top row) and ours (bottom row).}

    \label{fig:compare_multi_subject}
\end{figure}

\begin{figure}[t] 
    \centering


    \begin{minipage}{\linewidth}
        \scriptsize
        \raggedright
        \textbf{Prompt:} \textit{\textbf{Panda} is at a karaoke bar, singing loudly with his brothers, closing his eyes as if he were a superstar on TV.}
    \end{minipage}
    \includegraphics[width=\linewidth]{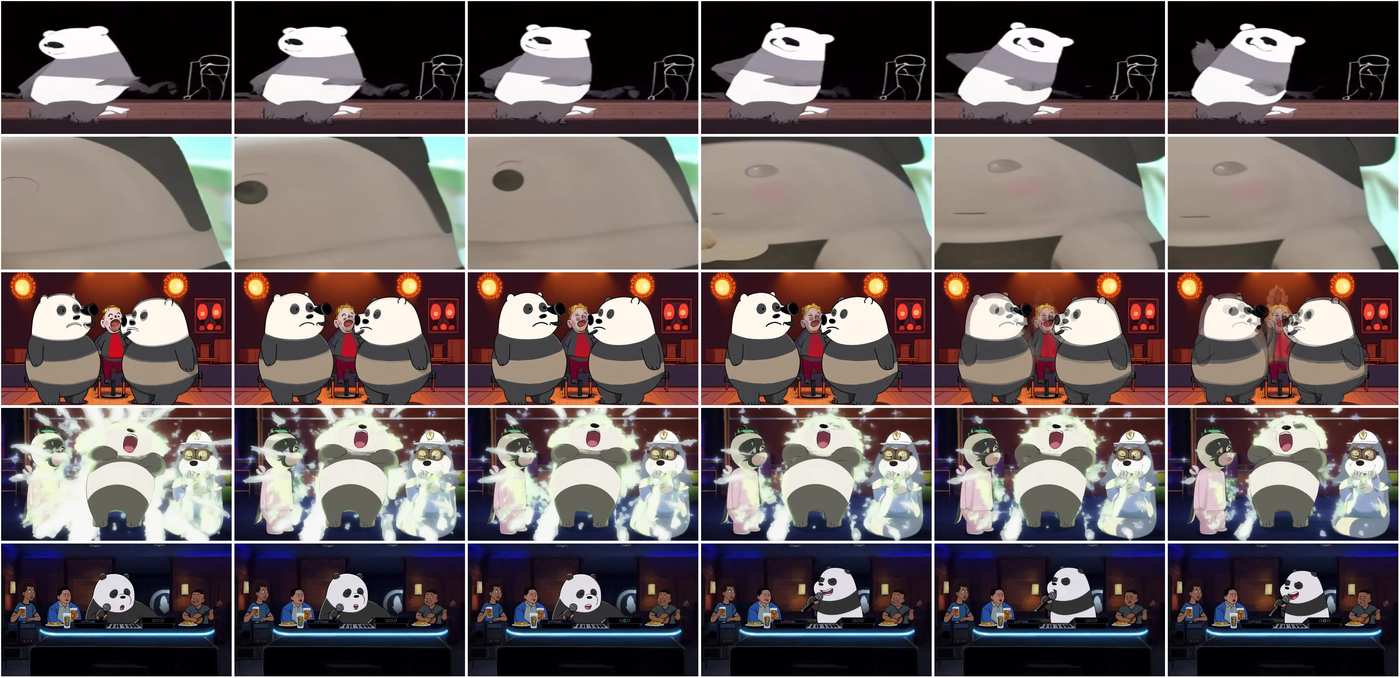}

    \begin{minipage}{\linewidth}
        \scriptsize
        \raggedright
        \textbf{Prompt:}  \textit{\textbf{Mr. Bean} is sitting alone on a park bench, trying to eat his sandwich while a bird keeps stealing the crumbs, making him look frustrated but also funny.}
    \end{minipage}
    \includegraphics[width=\linewidth]{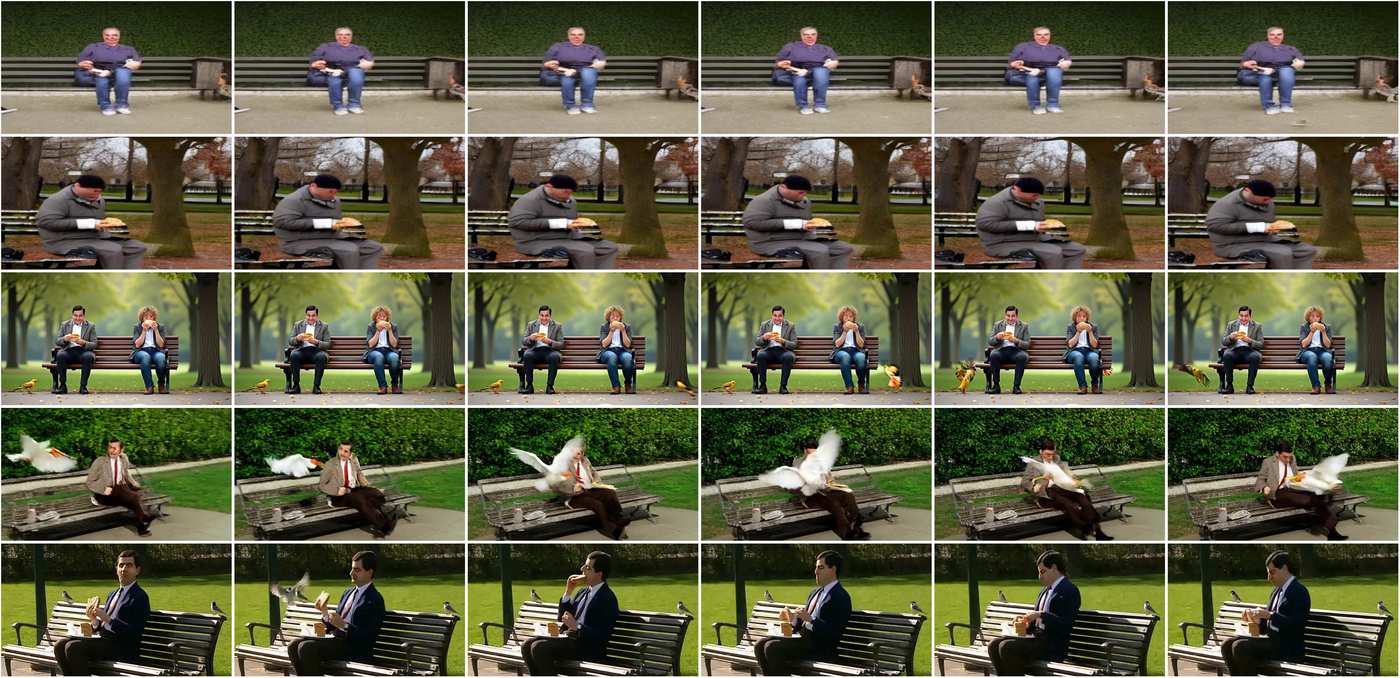}
     
    \caption{\textbf{Comparison on single-subject generation.} From top to bottom: results from VideoBooth~\cite{videobooth}, DreamVideo, Wan2.1-I2V~\cite{wan2025video}, SkyReel-A2~\cite{fei2025skyreelsa2} and ours.}

    \label{fig:compare_sinngle_subject}
\end{figure}

\section{Experiments}

\begin{figure}[t] 
    \centering
      
    \begin{minipage}{\linewidth}
        \scriptsize
        \raggedright
        \textbf{Prompt:} \textit{\textbf{Tom} plays piano loudly. \textbf{Jerry} dances on the keys. \textbf{Mr. Bean}, wearing earmuffs with his suit, tries to conduct them like an orchestra. 
        It turns into noisy chaos. The scene is cartoon style. }
    \end{minipage}
    \includegraphics[width=\linewidth]{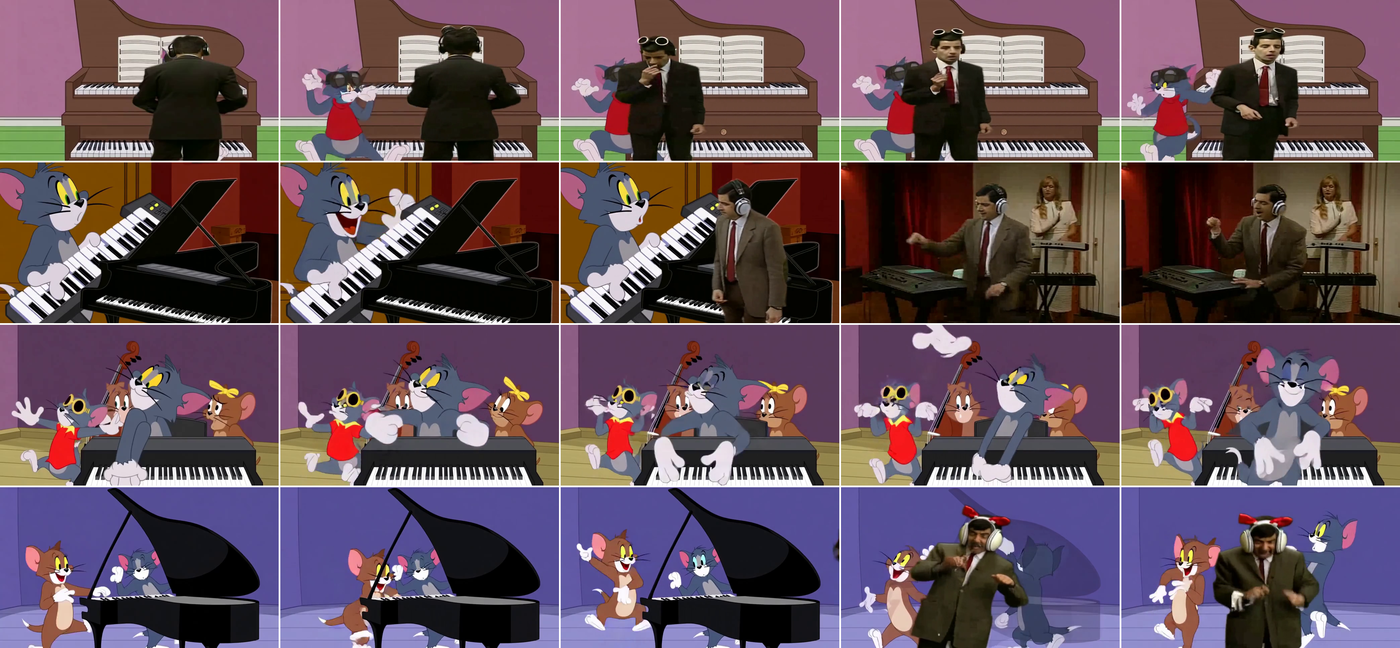}
      
    \caption{\textbf{Ablation on caption format.} From top to bottom: without tags, with \texttt{[character]} tag, with \texttt{[scene-style]} tag, and with both \texttt{character} and \texttt{scene-style} tags.}
    \label{fig:ablation_caption}
\end{figure}

\subsection{Benchmarks}
We evaluate our method using a comprehensive set of metrics along two dimensions. For overall video quality and temporal coherence, we adopt Consistency, Motion, Dynamic, Quality, and Aesthetic from VBench~\cite{vbench}. To assess character-level consistency and interaction, we employ a vision–language model (VLM) and introduce four specialized metrics: Identity-P, Motion-P, Style-P, and Interaction-P. Specifically, we leverage Gemini-1.5-Flash~\cite{gemini2024} as the VLM backbone for these evaluations. 

\qheading{Video Quality and Temporal Consistency.}
(1) \textbf{Consistency} evaluates the overall video-text consistency across frames computed by ViCLIP~\cite{wang2023internvid}. 
(2) \textbf{Motion} measures the level of smoothness of generated motions. 
(3) \textbf{Dynamic} quantifies the degree of motion dynamics using RAFT~\cite{raft}.  
(4) \textbf{Quality} measures the imaging quality, referring to the distortion (e.g., over-exposure, noise, blur) by the image quality predictor MUSIQ~\cite{ke2021musiq}.  
(5) \textbf{Aesthetic} evaluates the artistic and beauty value perceived by humans towards each video frame using the LAION aesthetic predictor.

\qheading{Character Consistency and Interaction.} 

\textbf{Identity-P} evaluates how well the generated video preserves each character’s visual identity and distinctive features. The VLM assesses facial feature consistency, body proportions, characteristic attributes (e.g., Jerry’s mouse ears, Tom’s whiskers), and overall color scheme. A score of 10 indicates perfect identity preservation, where the character is immediately recognizable; a score of 1 indicates the character is completely unrecognizable.

\textbf{Motion-P} measures the authenticity of character-specific movements and behaviors relative to their canonical personality traits. The evaluation considers motion patterns (e.g., Jerry’s quick scurrying, Tom’s exaggerated sneaking), behavioral consistency, and expression of personality through movement. The VLM analyzes temporal sequences to assess alignment with the character’s known behavior patterns.

\textbf{Style-P} assesses the consistency of each character’s original artistic and visual style. This includes animation style (e.g., cartoon vs. realistic), aesthetic coherence with the source material, art direction fidelity, and rendering consistency. The VLM compares the generated video to its learned representation of the character’s canonical appearance and stylistic conventions.  

\textbf{Interaction-P} evaluates the naturalness and plausibility of multi-character interactions. The assessment considers spatial relationships, timing and coordination, believability of reactions and responses, and physical dynamics between characters. For single-character videos, this metric evaluates interactions with the environment and scene elements.

\subsection{Comparison}
We benchmark our method against state-of-the-art video generation baselines, including two single-subject customization approaches—VideoBooth~\cite{videobooth} and DreamVideo~\cite{dreamvideo}—as well as foundation image-to-video models Wan2.1-I2V~\cite{wan2025video} and SkyReels-A2~\cite{fei2025skyreelsa2}, both of which support single- and multi-character generation. For multi-subject customization specifically, we compare directly against SkyReels-A2. 
Note that Wan2.1-I2V cannot directly generate videos from a character image. Thus, we first employ OmniGen~\cite{xiao2025omnigen} to synthesize an image using the prompt and reference, which is then used as input to Wan2.1-T2V.

For single-subject evaluation, we generate 50 videos featuring 10 characters—five from cartoons (\texttt{Tom}, \texttt{Jerry}, \texttt{Grizzly}, \texttt{Ice Bear}, \texttt{Panda}) and five from live-action series (\texttt{Mr. Bean}, \texttt{Sheldon}, \texttt{George}, \texttt{Mary}, \texttt{Penny}). 
For multi-subject evaluation, we generate 50 videos, each featuring 2–3 characters interacting within the same scene (noting that SkyReels-A2 supports fewer than three characters). These interactions span a wide range of scenarios, including inter-style (cartoon with real-life), intra-style (within cartoons or within real-life), inter-series (across different shows), and intra-series (within the same show). All reference images are included in the Appendix.

Figure~\ref{fig:compare_sinngle_subject} shows the qualitative comparison on single subject video generation.  
VideoBooth~\cite{videobooth} and DreamVideo~\cite{dreamvideo} fail to preserve the visual identity of the reference character.
SkyReel-A2~\cite{fei2025skyreelsa2} and Wan2.1-I2T~\cite{wan2025video} retain identity to some extent—though they struggle with facial details—but fail to synthesize character-faithful motions.
In contrast, our method consistently preserves visual fidelity and generates character-faithful motion.
 
Figure~\ref{fig:compare_multi_subject} presents qualitative comparisons on multi-character interaction, both within the same style and across different styles. While SkyReel-A2~\cite{fei2025skyreelsa2} can synthesize acceptable single-subject videos, it struggles with complex interactions across multiple characters, especially in multi-style settings. Although it can place several characters into a shared scene, the results often exhibit visual inconsistencies and unnatural interactions. In contrast, our method enables contextually coherent interactions without compromising character identity or native style.

Quantitative results across nine metrics are reported in Table~\ref{tab:comparison}. Our method consistently outperforms prior approaches in both single- and multi-subject settings, demonstrating stronger identity preservation, faithful motion synthesis, and coherent style maintenance across diverse interaction scenarios. Additional comparison results are provided in the Appendix.
 
\begin{figure}[t] 
    \centering

    \begin{minipage}{\linewidth}
        \scriptsize
        \raggedright
        \textbf{Prompt:} \textit{\textbf{Mr. Bean} rides a unicycle. \textbf{Ice Bear} juggles. \textbf{Panda} plays trumpet. The stage explodes fire.}
    \end{minipage}
    \includegraphics[width=\linewidth]{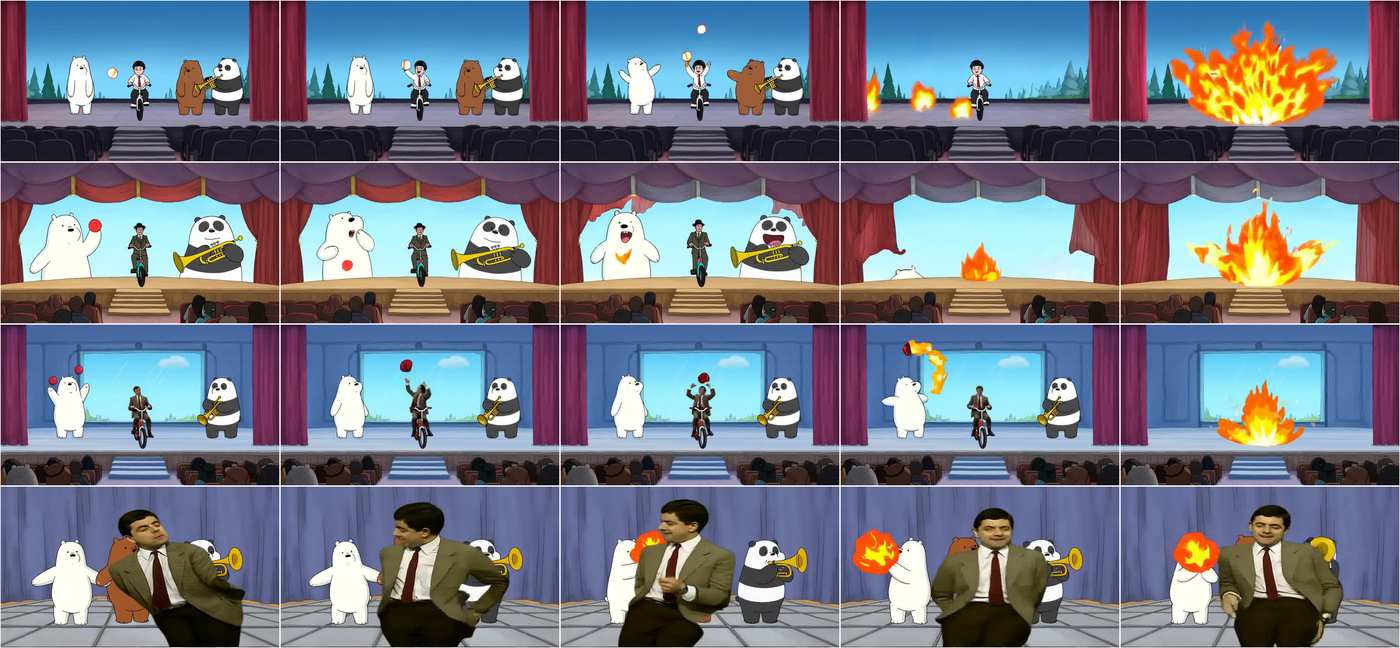}
    
    \caption{\textbf{Ablation on augmentation data ratio.} From top to bottom: 0\%, 5\%, 10\%, and 20\% augmentation.}
    \label{fig:ablation_dataratio}
\end{figure}

\begin{table*}[t!]
\centering
\caption{
\textbf{Comparison of recent video generation models across evaluation dimensions.} 
The first group of columns includes automatic evaluation metrics, while the last three report human evaluation scores.
\textbf{Bold} indicates the best performance per column. 
}
\vspace{-6pt}
\resizebox{1\linewidth}{!}{
\begin{tabular}{>{\centering\arraybackslash}p{5.5cm} >{\centering\arraybackslash}p{1.2cm} 
>{\centering\arraybackslash}p{2cm} >{\centering\arraybackslash}p{1.5cm} 
>{\centering\arraybackslash}p{1.5cm} >{\centering\arraybackslash}p{1.2cm} >{\centering\arraybackslash}p{1.4cm}
 >{\centering\arraybackslash}p{1.5cm} >{\centering\arraybackslash}p{1.4cm} >
 {\centering\arraybackslash}p{1.3cm} >
 {\centering\arraybackslash}p{2.0cm}
}
\rowcolor{headergray}
\specialrule{1.2pt}{0pt}{3pt}
& 
& 
\multicolumn{5}{c}{\textbf{VBench Metrics} \cite{vbench}} & 
\multicolumn{4}{c}{\textbf{VLM Metrics}} \\
 \cmidrule(lr){3-7} \cmidrule(lr){8-11}

 \rowcolor{headergray}
\textbf{Methods} &  \textbf{Subject} & 
Consistency &  
Motion & 
Dynamic & 
Quality  & 
Aesthetic  & 
Identity-P & 
Motion-P &
Style-P  & 
Interaction-P\\
\specialrule{0.8pt}{3pt}{3pt}

\rowcolor{baselineblue}
VideoBooth~\cite{videobooth}         & Single   & 0.1287    & 0.9780 & 0.5094 & 0.6413 & 0.4896  &  4.45 &  3.72 & 5.43 & 4.44   \\

\rowcolor{baselineblue}
DreamVideo~\cite{dreamo2024}      & Single   & 0.1851 &     0.9564 &   & 0.6270 &  0.5002  & 4.51 & 4.16 & 6.82 & 5.37  \\

\rowcolor{baselineblue} 
Wan2.1-I2V~\cite{wan2025video}   & Single   & 0.0682 &    0.9827 & 0.6530 & 0.7192 & 0.5857 & 5.27 & 5.10 & 7.94 & 6.41\\

\rowcolor{baselineblue}
SkyReels-A2~\cite{chen2025skyreelsv2}        & Single   &  0.1469      & 0.9782 & 0.7843 & 0.7225 & 0.5850 &6.17 & 4.55 & 7.82  & 6.78 \\

\rowcolor{highlightgold}
\textbf{Ours}   & Single   &  \textbf{0.1893} &     \textbf{0.9836}& \textbf{1.0000} & 0.5763 & \textbf{0.5967}  & \textbf{6.12} & \textbf{5.41} &  \textbf{8.06} &  \textbf{7.24}  \\

\specialrule{0.8pt}{3pt}{3pt}

\rowcolor{baselineblue}
SkyReels-A2~\cite{chen2025skyreelsv2}       & Multiple &  0.1314 &   0.9650 & 0.9787 & 0.7140   & 0.5371   &6.17 & 4.55 & 6.28& 4.94   \\

\rowcolor{highlightgold}
\textbf{Ours}        & Multiple & \textbf{0.1833}   & \textbf{0.9842  } & \textbf{0.98555} & 0.6855 & \textbf{0.5813} &  \textbf{6.48} & \textbf{5.50} & \textbf{7.26}&\textbf{5.22}  \\

\specialrule{1.2pt}{3pt}{0pt}
\end{tabular}
}
\label{tab:comparison}
\end{table*}

\begin{table*}[t!]
\centering 
\caption{
\textbf{Effect of Caption Formatting.} 
We compare different caption formats, with and without structured tags, across multiple evaluation dimensions. 
Our full formatting with both \texttt{[character]} and \texttt{[scene-style]} tags achieves the best performance.
}
\vspace{-6pt}
\resizebox{1\linewidth}{!}{
\begin{tabular}{>{\centering\arraybackslash}p{3.5cm}  
>{\centering\arraybackslash}p{1.5cm} >{\centering\arraybackslash}p{2.2cm} 
>{\centering\arraybackslash}p{1.5cm} >{\centering\arraybackslash}p{1.2cm} >{\centering\arraybackslash}p{1.4cm}
 >{\centering\arraybackslash}p{1.5cm} >{\centering\arraybackslash}p{1.4cm} >
 {\centering\arraybackslash}p{1.3cm} >
 {\centering\arraybackslash}p{2.0cm}
}
\rowcolor{headergray} 
\specialrule{1.2pt}{0pt}{3pt}
 &  
\multicolumn{5}{c}{\textbf{VBench Metrics} \cite{vbench}} & 
\multicolumn{4}{c}{\textbf{VLM Metrics}} \\
 \cmidrule(lr){2-6} \cmidrule(lr){7-10}

\rowcolor{headergray}
 \textbf{Caption Format} &  
Subject-C & 
Background-C &  
Motion-S & 
Dynamic & 
Quality  & 
Identity-P & 
Motion-P &
Style-P  & 
Interaction-P\\
\specialrule{0.8pt}{3pt}{3pt}

\rowcolor{baselineblue}
No Tag (Baseline)          & \textbf{0.8892} & \textbf{0.9136} & \textbf{0.9812} &
  \textbf{1.0000} & 0.6535  &7.02 & 5.90 & 6.83 & 4.28    \\

\rowcolor{baselineblue}
w/o \texttt{[scene-style]}   &  0.8668 & 0.8980 & 0.9754 & \textbf{1.0000} & 0.6353  & 7.31 & 5.80 & \textbf{6.95} & 4.47    \\
 
\rowcolor{baselineblue} 
w/o \texttt{[character]}  & 0.8758 & 0.9101 & 0.9759 & \textbf{1.0000} & \textbf{0.6938}    &7.33 & 5.42 & 6.80 &   4.47  \\ 

\rowcolor{highlightgold}
w/ Both (Ours)   &  0.8530 & 0.8997 & 0.9747 & \textbf{1.0000} & 0.6588 &  \textbf{7.35} & 5.80 & \textbf{6.95} &   \textbf{5.30}  \\

\specialrule{1.2pt}{3pt}{0pt}
\end{tabular}
}
\label{tab:caption_format}
\end{table*}

\begin{table*}[t!]
\centering 
\caption{
\textbf{Effect of Synthetic Data Augmentation.}
We vary the proportion of synthetic videos relative to the original dataset and evaluate across the same dimensions used in Table~\ref{tab:comparison}.
}
\vspace{-6pt}
\resizebox{1\linewidth}{!}{
\begin{tabular}{>{\centering\arraybackslash}p{2cm}  
>{\centering\arraybackslash}p{1.5cm} >{\centering\arraybackslash}p{2.2cm} 
>{\centering\arraybackslash}p{1.5cm} >{\centering\arraybackslash}p{1.2cm} >{\centering\arraybackslash}p{1.4cm}
 >{\centering\arraybackslash}p{1.5cm} >{\centering\arraybackslash}p{1.4cm} >
 {\centering\arraybackslash}p{1.3cm} >
 {\centering\arraybackslash}p{2.0cm}
}
\rowcolor{headergray} 
\specialrule{1.2pt}{0pt}{3pt}
 &  
\multicolumn{5}{c}{\textbf{VBench Metrics} \cite{vbench}} & 
\multicolumn{4}{c}{\textbf{VLM Metrics}} \\
 \cmidrule(lr){2-6} \cmidrule(lr){7-10}

\rowcolor{headergray}
 \textbf{Ratio} &  
Subject-C & 
Background-C &  
Motion-S & 
Dynamic & 
Quality  & 
Identity-P & 
Motion-P &
Style-P  & 
Interaction-P\\
\specialrule{0.8pt}{3pt}{3pt}

\rowcolor{baselineblue}
\rowcolor{baselineblue}
5\%      &  0.8739 & 0.9082 & 0.9826 & 0.9000 & 0.6836 &  7.67 &  6.03 & 7.15 & \textbf{4.83}   \\

\rowcolor{highlightgold} 
10\%   &   \textbf{0.8812} & \textbf{0.9151} & \textbf{ 0.9853} & 0.9500 & \textbf{0.6955}  & \textbf{8.33} & \textbf{6.72} & \textbf{7.33} & 4.78   \\

\rowcolor{baselineblue} 
20\% &   0.8442 & 0.8905 & 0.9779 & \textbf{1.0000}& 0.6728 & 8.30 & 7.10 & 7.08 & 3.90   \\

\specialrule{1.2pt}{3pt}{0pt}
\end{tabular}
}
\label{tab:augratio}
\end{table*}


\subsection{Ablation study} 
We conduct ablation studies to investigate how different captioning strategies influence the model’s ability to learn and ground each character as a distinct concept. Additionally, we analyze how varying the ratio of augmentation data affects the mitigation of style delusion. 

\qheading{Captions Formats.}
To evaluate the role of structured captions, we compare models trained with standard free-form captions against those trained with captions augmented by our proposed tags \texttt{[scene-style]} and \texttt{[character]}. The structured format provides explicit grounding of both scene attributes and character identities. Figure~\ref{fig:ablation_caption} illustrates a representative example of our video–caption pairs, showing how the tags enable more faithful alignment between visual entities and textual descriptions. The second row (without the \texttt{[scene-style]} tag) shows a shift from a cartoon scene to a realistic one. The third row (without the \texttt{[character]} tag) fails to include Mr. Bean in the generated video. In contrast, our method, using both tags, produces consistent scenes and correctly includes all characters.

\qheading{Augment Data Ratio.} 
We analyze the effect of incorporating our composited dataset (Section~\ref{sec:augmentation}) at different mixing ratios. Specifically, we vary the proportion of synthetic videos relative to the original curated dataset and evaluate the impact on performance. This experiment highlights how synthetic cross-character interactions contribute to improved generalization and robustness in multi-character video generation.
Figure~\ref{fig:ablation_dataratio} illustrates a representative example. Both the baseline (0\% augmentation) and the 5\% augmentation setting generate a cartoon-style scene. While the 5\% setting correctly produces a cartoon-style Mr. Bean, the 0\% setting fails to preserve identity, instead generating a random cartoon character—possibly resembling one from the \threebears,series. At 10\% augmentation, the model successfully generates a realistic Mr. Bean with plausible interaction within the scene. However, at 20\%, although Mr. Bean’s appearance remains realistic, the interaction becomes less coherent, likely due to the overuse of synthetic data.



\section{Discussion}
Despite the effectiveness of our framework in enabling controllable multi-character video generation, it comes with several limitations. Most notably, our approach relies on explicit identity annotations and LoRA fine-tuning. As a result, introducing a new character—whether from a different show or an unseen domain—requires retraining or fine-tuning the model. This limits scalability in open-world settings, where users may wish to generate videos with arbitrary or user-defined characters.

Furthermore, while our captioning and augmentation strategies mitigate style delusion and enable robust character disentanglement, the model still exhibits occasional failure cases in highly complex interaction scenes, especially when multiple characters with overlapping appearances or motion patterns are present. 

\bibliography{main}
\bibliographystyle{style/iclr2026_conference}

\appendix
\clearpage
\begin{center}
{\fontsize{22pt}{28pt}\selectfont Supplementary Material}
\end{center}

\section{Use of LLM}
We used a large language model (GPT-4) to assist with grammar correction and language refinement throughout the writing process. Specifically, the model was employed to polish sentence structure, improve clarity, and ensure consistency in technical terminology. No content or experimental results were generated by the model; all scientific contributions and analysis were conducted by the authors.

\section{Dataset}
\qheading{Character Distribution.} Based on the character distribution analysis across 52,792 video clips from four TV series, there is significant variation in character
  prominence. Sheldon dominates the \youngsheldon dataset, appearing in 67.2\% of the 24,447 clips, making him the most frequently featured
  character overall with 16,440 occurrences. The \threebears shows more balanced character distribution, with Grizzly (57.9\%), Panda
  (52.8\%), and Ice Bear (46.1\%) all appearing in roughly half of their 16,762 clips. Mr. Bean demonstrates the highest character
  consistency, appearing in 94.7\% of his show's 6,662 clips, reflecting the show's single-protagonist nature. In contrast, \tomandjerry
  exhibits a hierarchical character presence, with the titular characters Tom (70.9\%) and Jerry (48.8\%) appearing much more frequently than
   supporting characters like Spike (22.6\%), Tuffy (12.7\%), and Quacker (8.0\%). The \youngsheldon series shows the most diverse character
  representation with eight tracked characters, though secondary characters like Mary Cooper (29.8\%) and George Cooper (19.5\%) appear
  significantly less than the protagonist. Overall, the data reveals that protagonist-centered shows (\mrbean, \youngsheldon) maintain
  high main character presence, while ensemble shows (\threebears) and classic animation (\tomandjerry) display more varied character
  distribution patterns.
 
\begin{figure}[b!]
    \centering
    \includegraphics[width=0.9\linewidth]{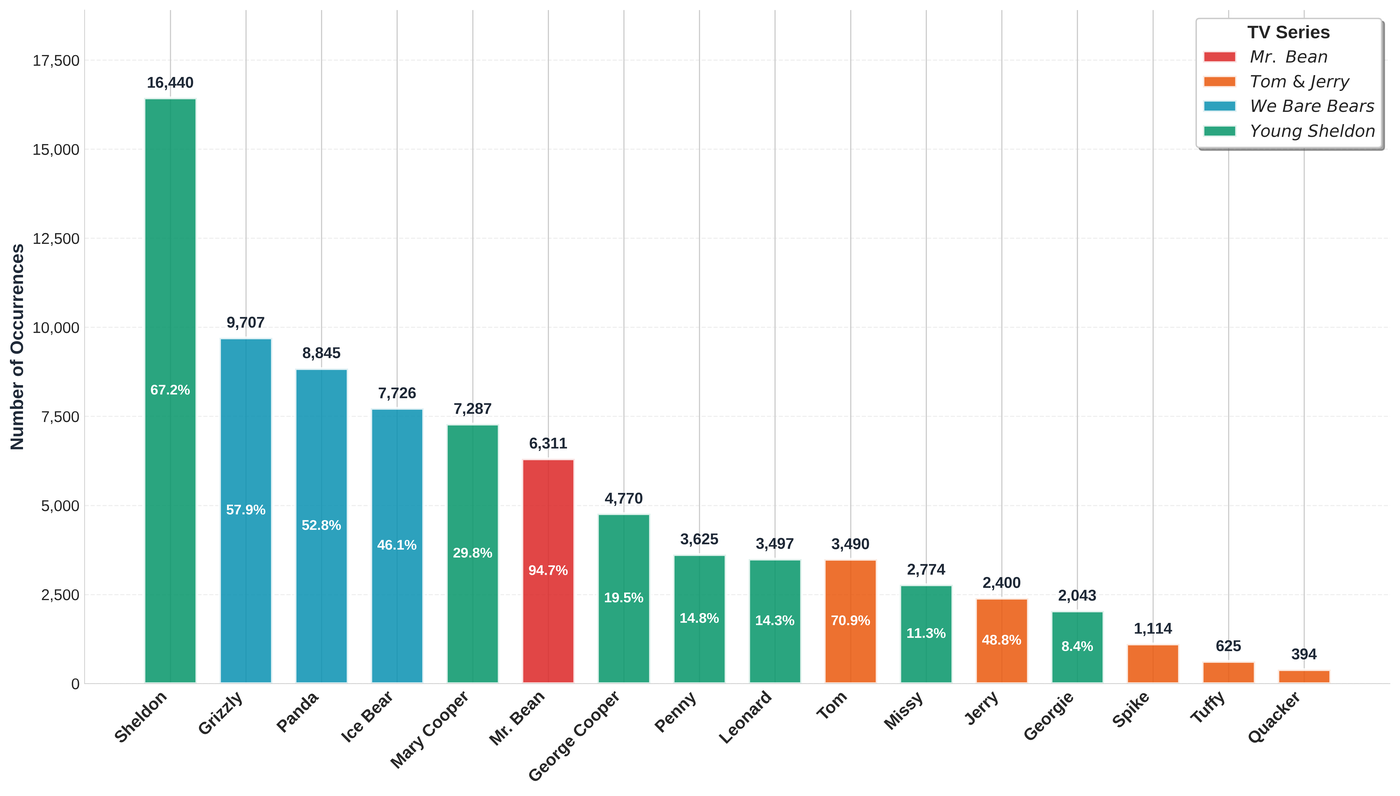}
    \caption{\textbf{Character distribution in our dataset.} 
    The bar chart shows the number of appearances of each character in the video-caption pairs.}
    \label{fig:character_distribution}
\end{figure}

\qheading{Captioning Prompt to GPT-4o.} 
We provide GPT-4o with 10 frames and the corresponding transcript, along with carefully designed instructions to ensure precise and consistent character references. Prompts for the four shows are shown below. 

\begin{tcolorbox}[title=\mrbean]
The video is from the comedy series Mr. Bean. 

If multiple people appear, please refer to him as 'Mr. Bean' and make it clear who he is. 

Do not use vague terms like 'a man' or 'someone' - always specify Mr. Bean for the main subject. 

Do not mention the title or say 'this video is from Mr. Bean'. 

Avoid phrases like 'seems to', 'suggests', 'may', or 'appears'.
\end{tcolorbox}

\begin{tcolorbox}[title=\tomandjerry]
This video clip is from the classic cartoon Tom and Jerry. 

The main characters are:

- Tom: a gray cat who usually chases Jerry

- Jerry: a small brown mouse who outsmarts Tom

- Spike: a gray dog who is Tom's enemy

- Tuffy: a small white mouse who is Jerry's friend

- Quacker: a yellow duck who is Tom's rival

Always refer to them by their names 'Tom', 'Jerry', 'Spike', 'Tuffy', 'Quacker'. 

Do not use vague terms like 'a cat', 'a mouse', or 'someone'. 

Make it clear who is doing what in the video. 

Do not mention the show's title or say 'this video is from Tom and Jerry'. 

Avoid phrases like 'seems to', 'suggests', 'may', or 'appears'.
\end{tcolorbox}

\begin{tcolorbox}[title=\threebears]
This video clip is from the cartoon We Bare Bears, which features three main characters:

- Grizzly (or Grizz): a brown bear who is outgoing and energetic

- Panda: a black-and-white panda bear who is shy and loves technology

- Ice Bear: a white polar bear who speaks in third person and is very calm and skilled

If multiple bears appear in the video, always refer to them by their character names: 

'Grizzly', 'Panda', or 'Ice Bear'. Do not use vague terms like 'a bear', 'a white bear', or 'someone'. 

Make it clear who is doing what in the video. Your job is to describe what happens in the video clip in a clear and detailed way using these character identities.  

Do not mention the show's title or say 'this video is from We Bare Bears'. 

Avoid phrases like 'seems to', 'suggests', 'may', or 'indicating'. 
\end{tcolorbox}

\begin{tcolorbox}[title=\youngsheldon]
This video clip is from the comedy series Young Sheldon, which features three main characters:

- Sheldon: a young boy who is very smart and has a unique way of thinking

- Penny: a girl who is Sheldon's best friend and has a crush on him

- Leonard: a boy who is Sheldon's best friend and has a crush on Penny

- Mary Cooper: a woman who is Sheldon's mother

- George Cooper: a man who is Sheldon's father

- Georgie: a boy who is Sheldon's younger brother

- Missy: a girl who is Sheldon's younger sister  

If multiple people appear, please refer to them as 'Sheldon', 'Penny', or 'Leonard'. Do not use vague terms like 'a boy', 'a girl', or 'someone'. 

Make it clear who is doing what in the video. Your job is to describe what happens in the video clip in a clear and detailed way using these character identities.  
\end{tcolorbox}

\begin{figure}[t] 
    \centering 
    \includegraphics[width=\linewidth]{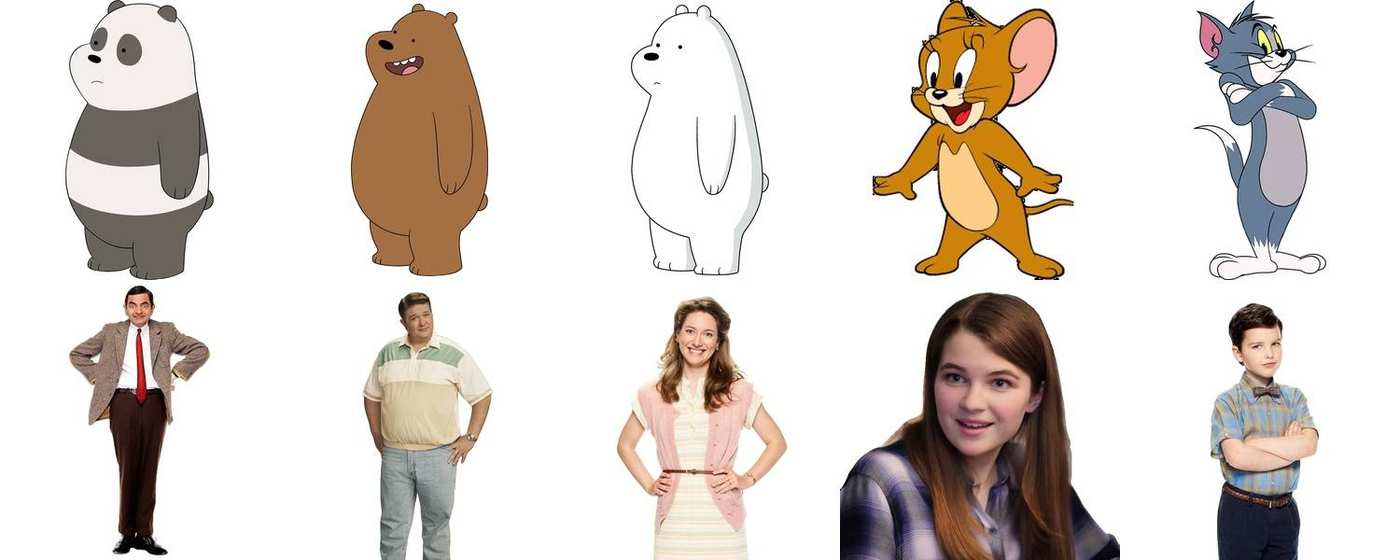} 
    \caption{\textbf{Reference images} of 5 cartoon characters including Panda, Grizzly, Ice Bear, Tom and Jerry, and 5 realistic characters including Mr. Bean, George Cooper, Mary Cooper, Penny and Sheldon.  } 

    \label{fig:supp_reference_image}
\end{figure}

\begin{figure}[t] 
    \centering

    \begin{minipage}{\linewidth}
        \scriptsize
        \raggedright
        \textbf{Prompt:} Panda, Mr. Bean, and Sheldon attend a tech startup pitch event. Panda pitches an app for bear selfies, Bean presents a banana phone, and Sheldon critiques everyone's latency stats. Nobody wins.
    \end{minipage}
    \includegraphics[width=\linewidth]{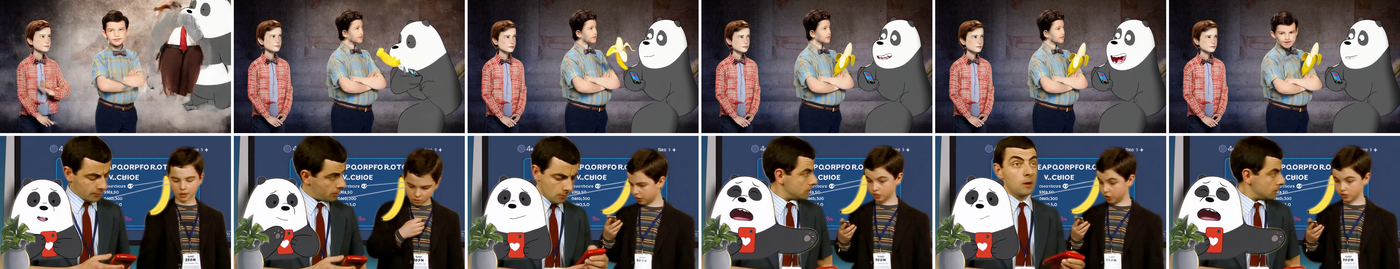}
    
    \begin{minipage}{\linewidth}
        \scriptsize
        \raggedright
        \textbf{Prompt:} Mr. Bean and Tom accidentally get locked inside a fancy hotel kitchen. While Tom chases a mouse under silver trays, Bean tries to cook pasta but ends up flooding the entire room with foam from a dishwasher he mistook for an oven.
    \end{minipage}
    \includegraphics[width=\linewidth]{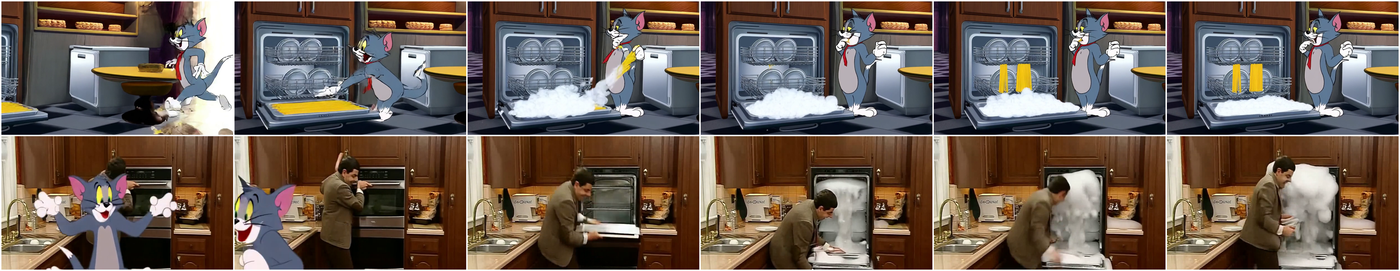}

    \begin{minipage}{\linewidth}
        \scriptsize
        \raggedright
        \textbf{Prompt:} Tom and Grizzly get jobs as mall security. Grizzly tries to be friendly with everyone, while Tom gets distracted chasing Jerry through every store. By the end, they both get promoted for “enthusiasm.”
    \end{minipage}
    \includegraphics[width=\linewidth]{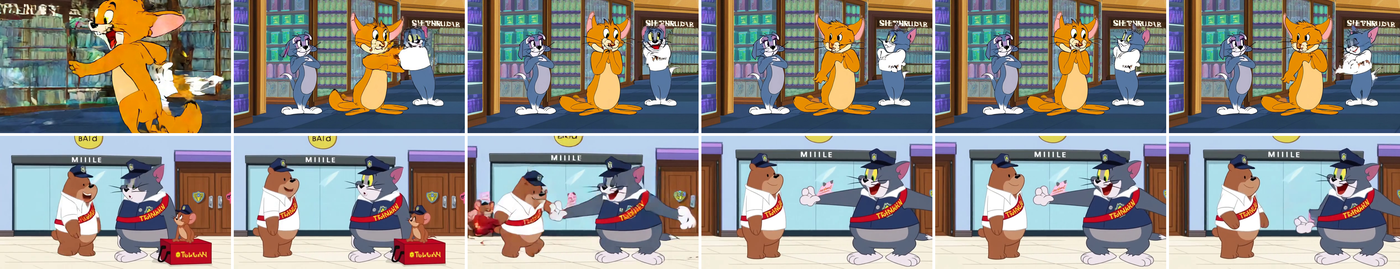}
    \caption{\textbf{Comparison on multi-subject interaction.} Results from SkyReel-A2~\cite{fei2025skyreelsa2} (top row) and ours (bottom row).}

    \label{fig:supp_compare_multi_subject}
\end{figure}


\section{Experiments}
\qheading{Reference images.} We provide 10 reference images together with their corresponding masks for evaluating VideoBooth~\cite{videobooth}, DreamVideo~\cite{dreamvideo}, Wan2.1~\cite{wan2025video}, and SkyReel-A2~\cite{fei2025skyreelsa2}, as illustrated in Figure~\ref{fig:supp_reference_image}.

\begin{figure}[t] 
    \centering 
    
    \begin{minipage}{0.95\linewidth}
        \scriptsize
        \raggedright
        \textbf{Prompt:} Tom is sitting on the floor with a bowl of milk, drinking it happily while closing his eyes in comfort.
    \end{minipage}
    \includegraphics[width=0.95\linewidth]{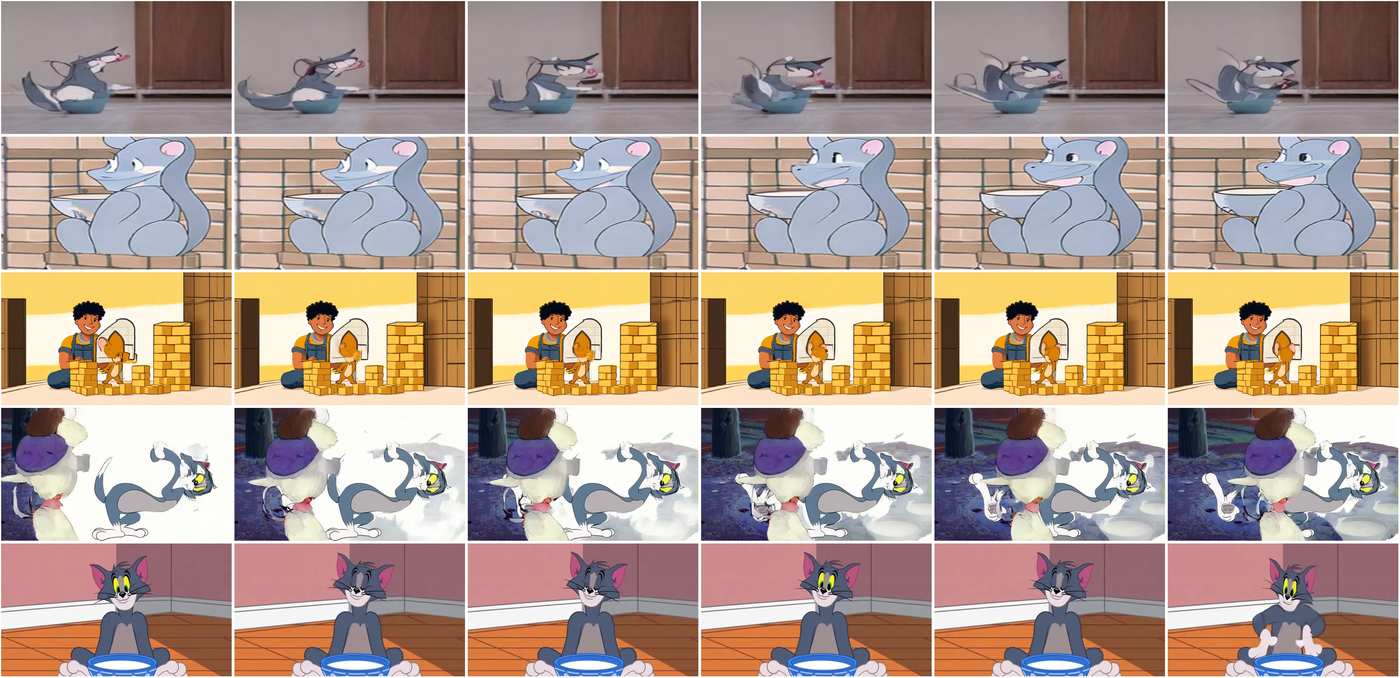}

    \begin{minipage}{0.95\linewidth}
        \scriptsize
        \raggedright
        \textbf{Prompt:} Grizzly is standing in a shopping mall, handing out flyers with too much enthusiasm, while people look confused but also amused by his energy.
    \end{minipage}
    \includegraphics[width=0.95\linewidth]{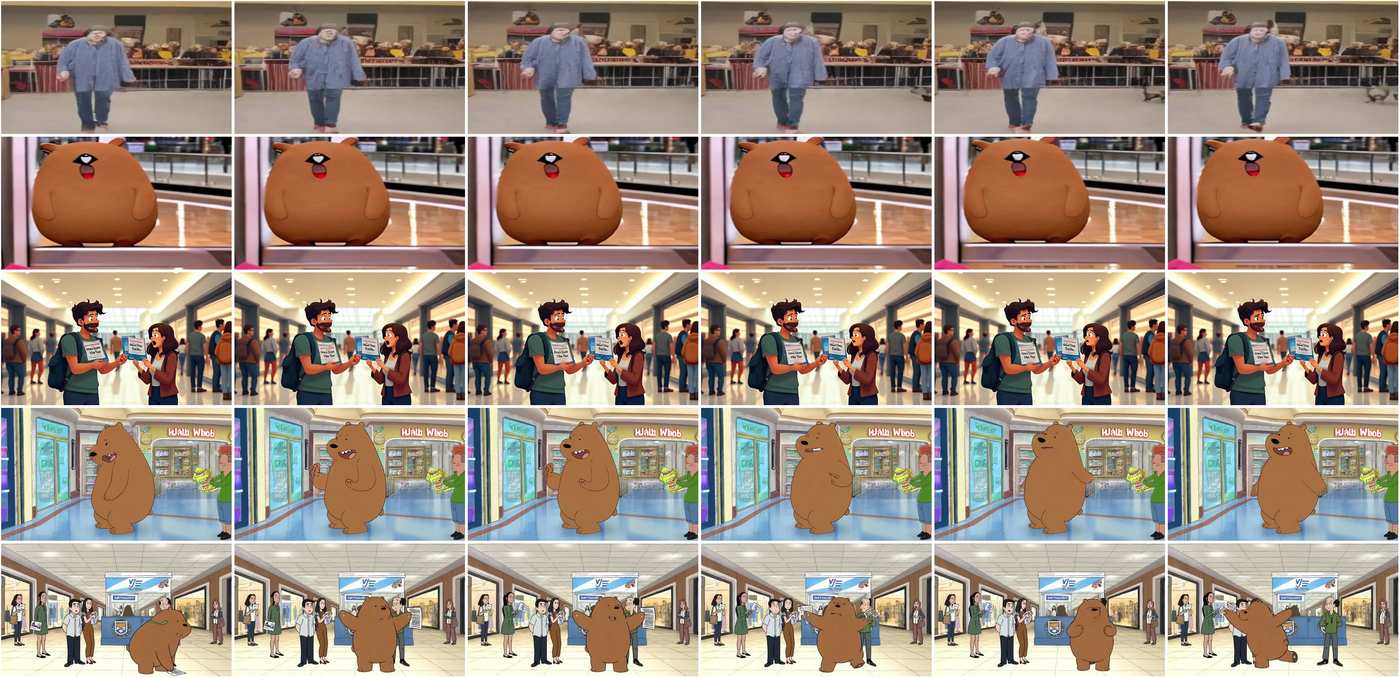}

    \begin{minipage}{0.95\linewidth}
        \scriptsize
        \raggedright
        \textbf{Prompt:} Ice Bear is silently cleaning the kitchen with great focus, wearing a small apron, showing his calm and serious personality compared to his brothers.
    \end{minipage}
    \includegraphics[width=0.95\linewidth]{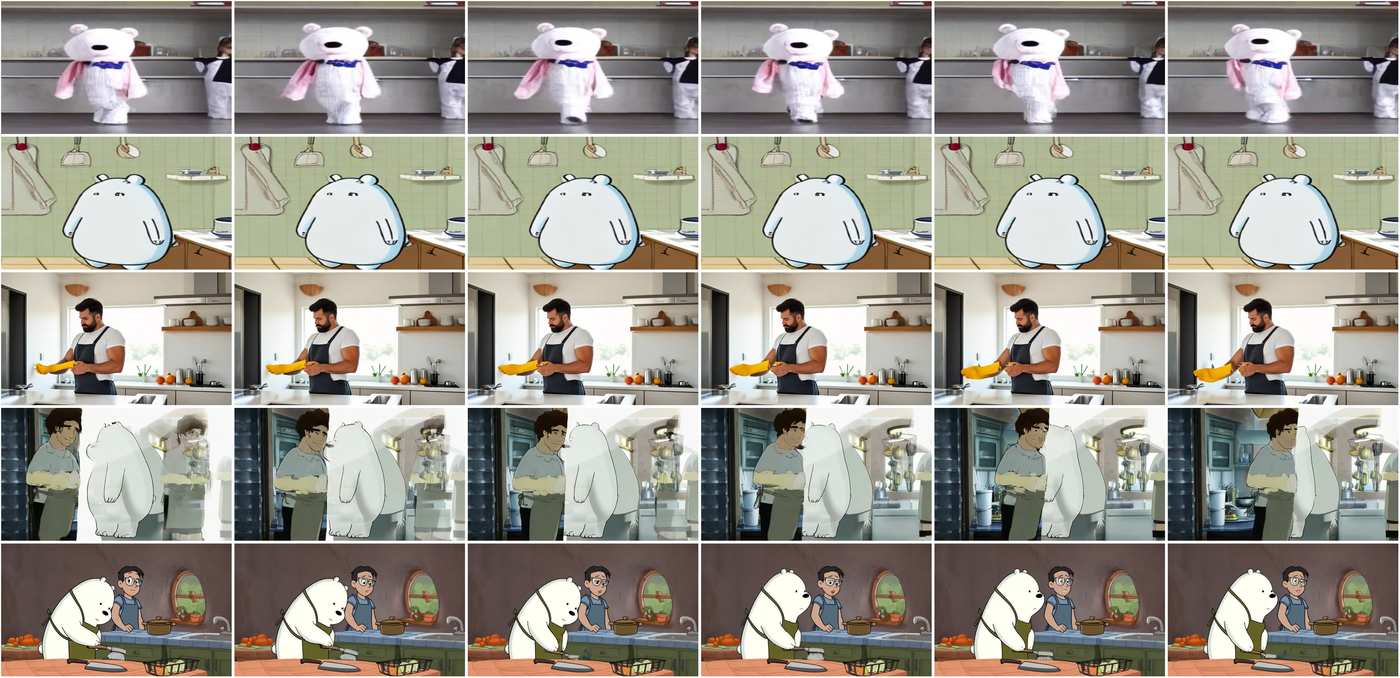}
   
    \caption{\textbf{Comparison on single-subject generation.} From top to bottom: results from VideoBooth~\cite{videobooth}, DreamVideo, Wan2.1~\cite{wan2025video}, SkyReel-A2~\cite{fei2025skyreelsa2} and ours.}

    \label{fig:supp_compare_sinngle_subject1}
\end{figure}

\begin{figure}[t] 
    \centering 
    
    \begin{minipage}{0.95\linewidth}
        \scriptsize
        \raggedright
        \textbf{Prompt:} Jerry is lying on a small bed made of cotton, sleeping peacefully with a little blanket covering him.
    \end{minipage}
    \includegraphics[width=0.95\linewidth]{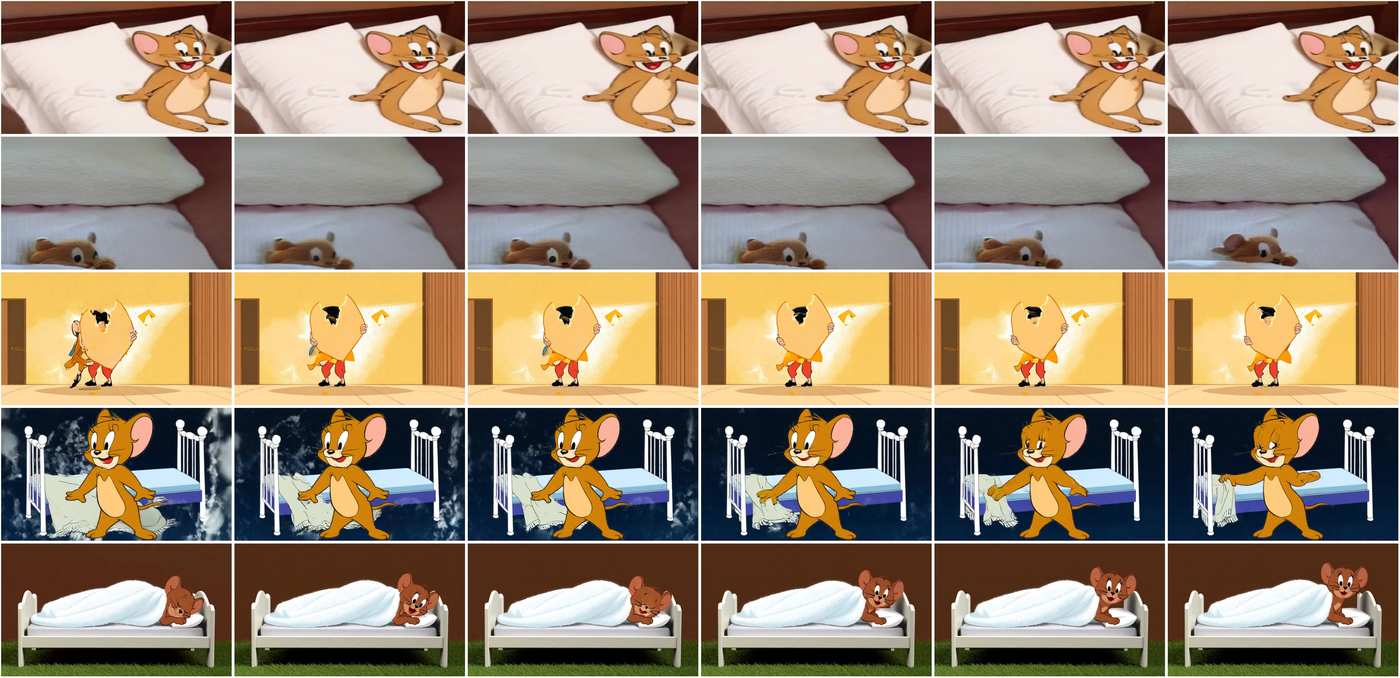}

    \begin{minipage}{0.95\linewidth}
        \scriptsize
        \raggedright
        \textbf{Prompt:} Ice Bear is sitting on the floor, drawing cute animal pictures in a sketchbook, keeping his talent hidden while everyone else is busy.
    \end{minipage}
    \includegraphics[width=0.95\linewidth]{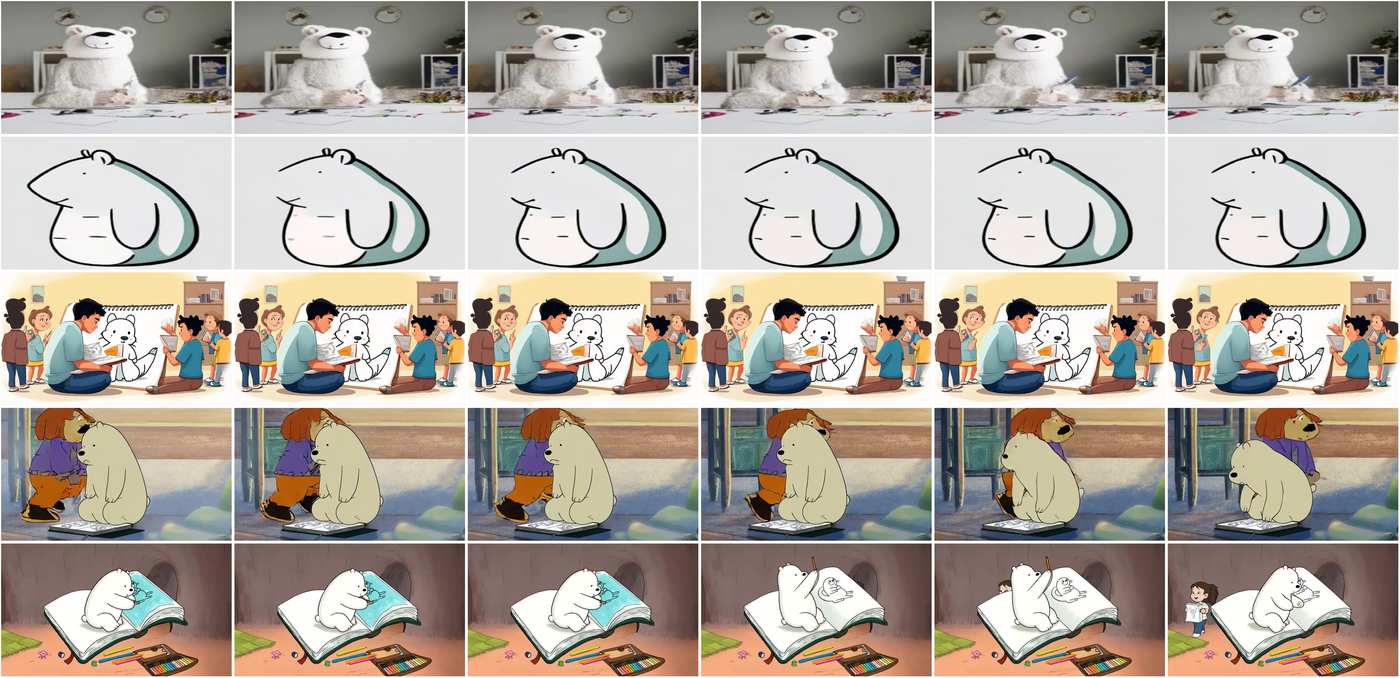}

    \begin{minipage}{0.95\linewidth}
        \scriptsize
        \raggedright
        \textbf{Prompt:} Panda is holding a big milk tea, sipping happily while scrolling through his favorite social media app, completely ignoring the world around him.
    \end{minipage}
    \includegraphics[width=0.95\linewidth]{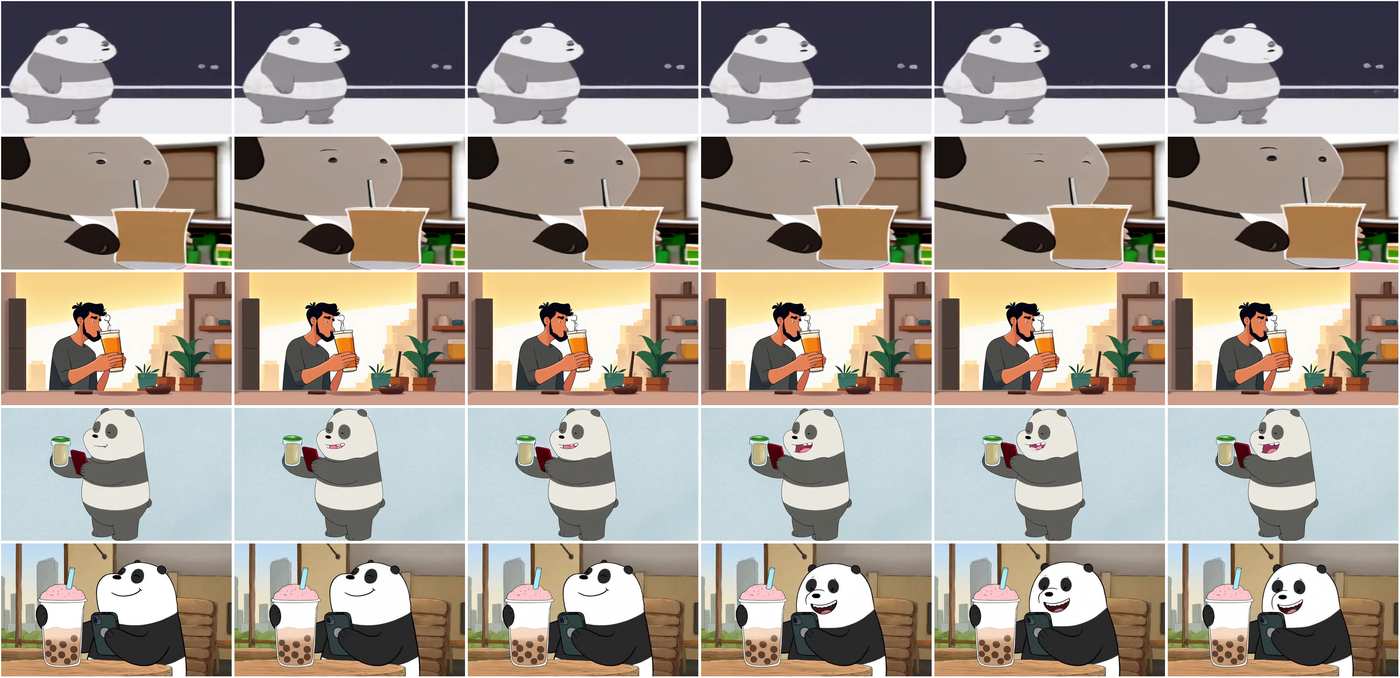}
   
    \caption{\textbf{Comparison on single-subject generation.} From top to bottom: results from VideoBooth~\cite{videobooth}, DreamVideo, Wan2.1~\cite{wan2025video}, SkyReel-A2~\cite{fei2025skyreelsa2} and ours.}

    \label{fig:supp_compare_sinngle_subject2}
\end{figure}

\begin{figure}[t] 
    \centering 
    

    \begin{minipage}{0.95\linewidth}
        \scriptsize
        \raggedright
        \textbf{Prompt:} Mr. Bean is sitting alone on a park bench, trying to eat his sandwich while a bird keeps stealing the crumbs, making him look frustrated but also funny.
    \end{minipage}
    \includegraphics[width=0.95\linewidth]{figure/images/comparison/01_mrbean_eat.jpg}
    
    \begin{minipage}{0.95\linewidth}
        \scriptsize
        \raggedright
        \textbf{Prompt:} Mr. Bean is in a supermarket, secretly opening candy boxes and tasting them before quickly putting them back, hoping no one notices his silly trick.
    \end{minipage}
    \includegraphics[width=0.95\linewidth]{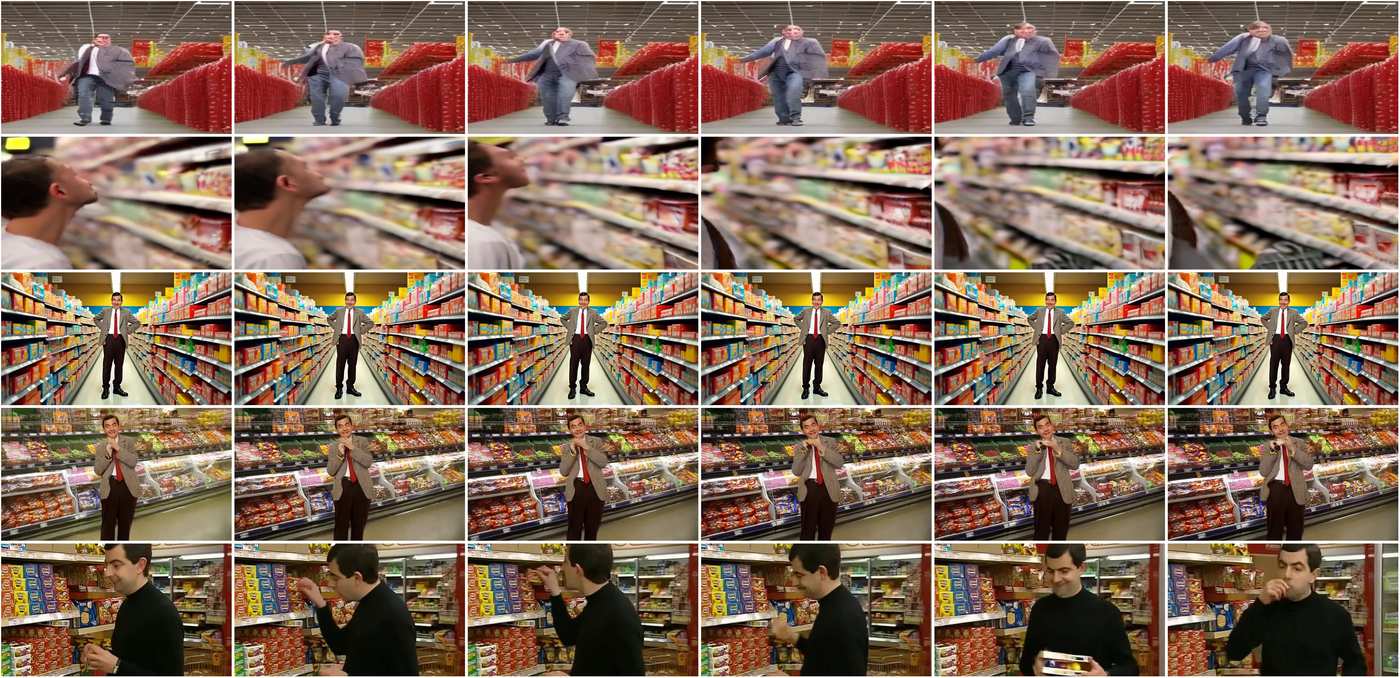}

    \begin{minipage}{0.95\linewidth}
        \scriptsize
        \raggedright
        \textbf{Prompt:} Mr. Bean is standing in front of a mirror, practicing strange faces and silly dance moves, enjoying his own performance as if it were a comedy show.
    \end{minipage}
    \includegraphics[width=0.95\linewidth]{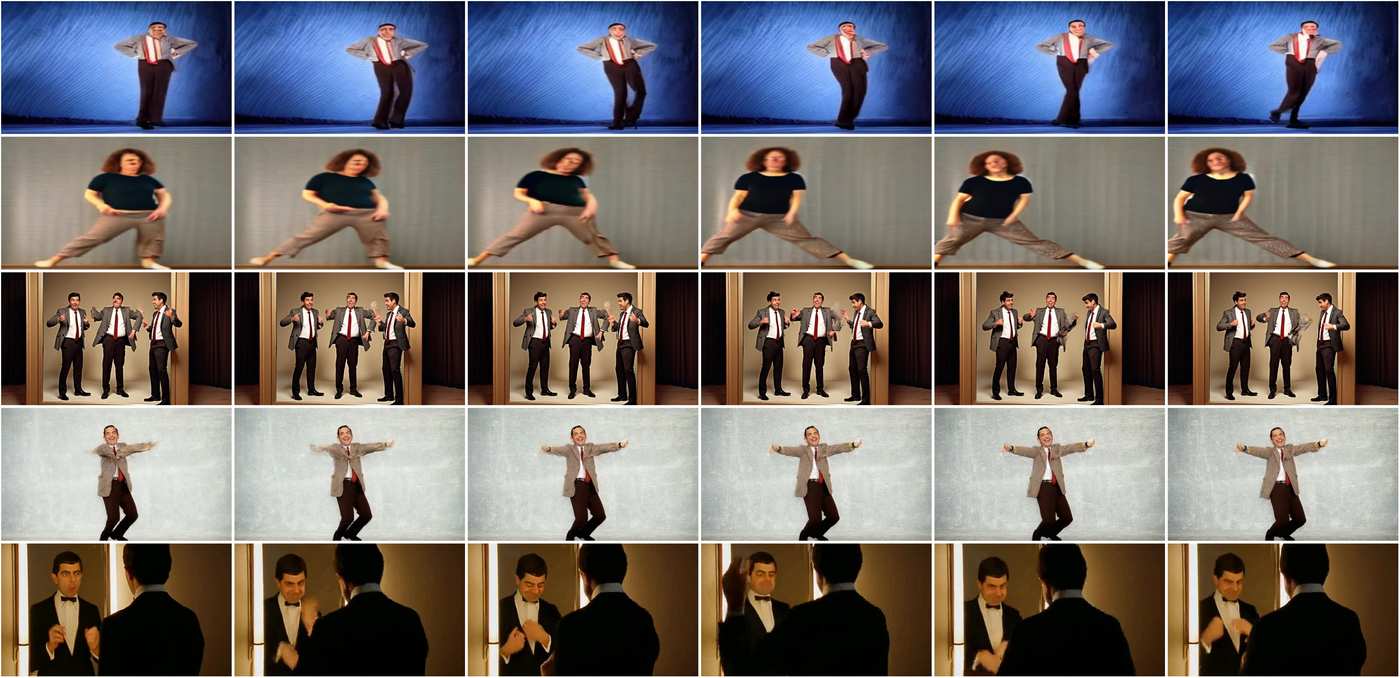}

    \caption{\textbf{Comparison on single-subject generation.} From top to bottom: results from VideoBooth~\cite{videobooth}, DreamVideo, Wan2.1~\cite{wan2025video}, SkyReel-A2~\cite{fei2025skyreelsa2} and ours.}

    \label{fig:supp_compare_sinngle_subject3}
\end{figure}

\qheading{Single-Subject Comparison.} We provide additional qualitative comparisons with VideoBooth~\cite{videobooth}, DreamVideo~\cite{dreamvideo}, Wan2.1~\cite{wan2025video}, and SkyReel-A2~\cite{fei2025skyreelsa2} in Figure~\ref{fig:supp_compare_sinngle_subject1},~\ref{fig:supp_compare_sinngle_subject2}, and~\ref{fig:supp_compare_sinngle_subject3}. These examples further demonstrate our method’s ability to preserve subject identity and produce coherent motion across diverse scenarios.

\qheading{Multi-Subject Comparison.} We present additional qualitative comparisons with SkyReel-A2~\cite{fei2025skyreelsa2} in Figure~\ref{fig:supp_compare_multi_subject}. The results demonstrate our method's capability to synthesize natural and coherent interactions among multiple characters while faithfully preserving their identities, characteristic behaviors, and distinctive visual styles. In comparison to SkyReel-A2, our approach yields fewer style inconsistencies and artifacts, producing more realistic and engaging multi-subject generations.

\begin{figure}[h] 
    \centering

    \begin{minipage}{0.9\linewidth}
        \scriptsize
        \raggedright
        \textbf{Prompt:}  Mary Cooper is planting flowers in the garden, wearing gloves and a hat, enjoying the sunshine and the quiet moment of peace.
    \end{minipage}
    \includegraphics[width=0.9\linewidth]{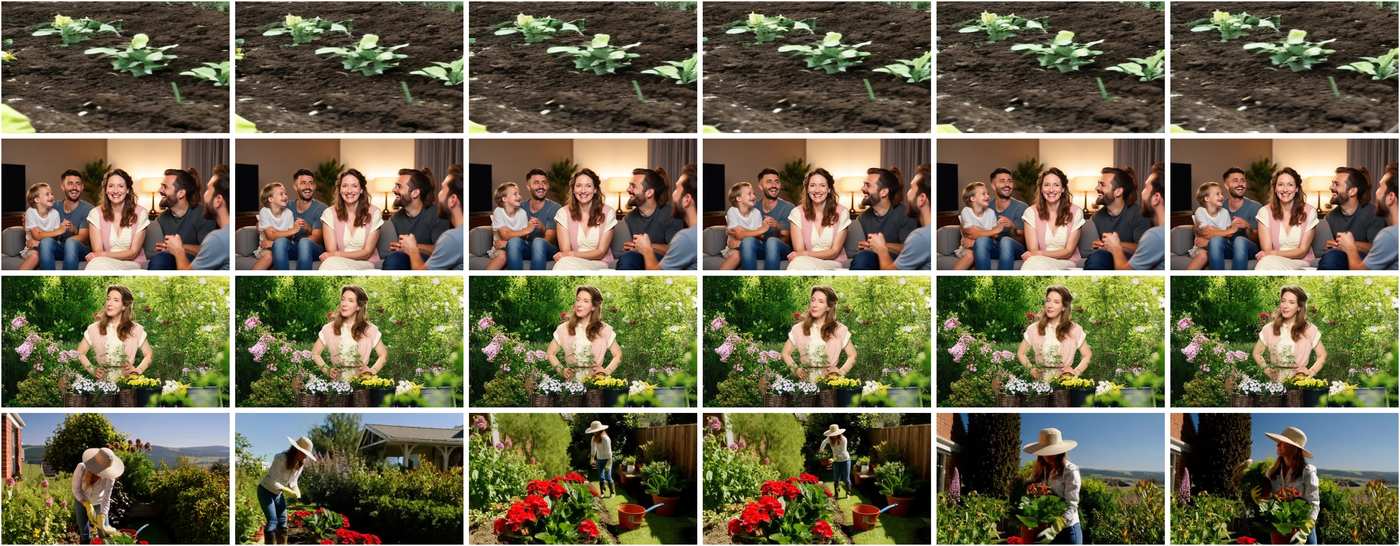}
    
    \begin{minipage}{0.9\linewidth}
        \scriptsize
        \raggedright
        \textbf{Prompt:}  George Cooper is barbecuing outside, flipping burgers with a big smile, while the family gathers around waiting for food.
    \end{minipage}
    \includegraphics[width=0.9\linewidth]{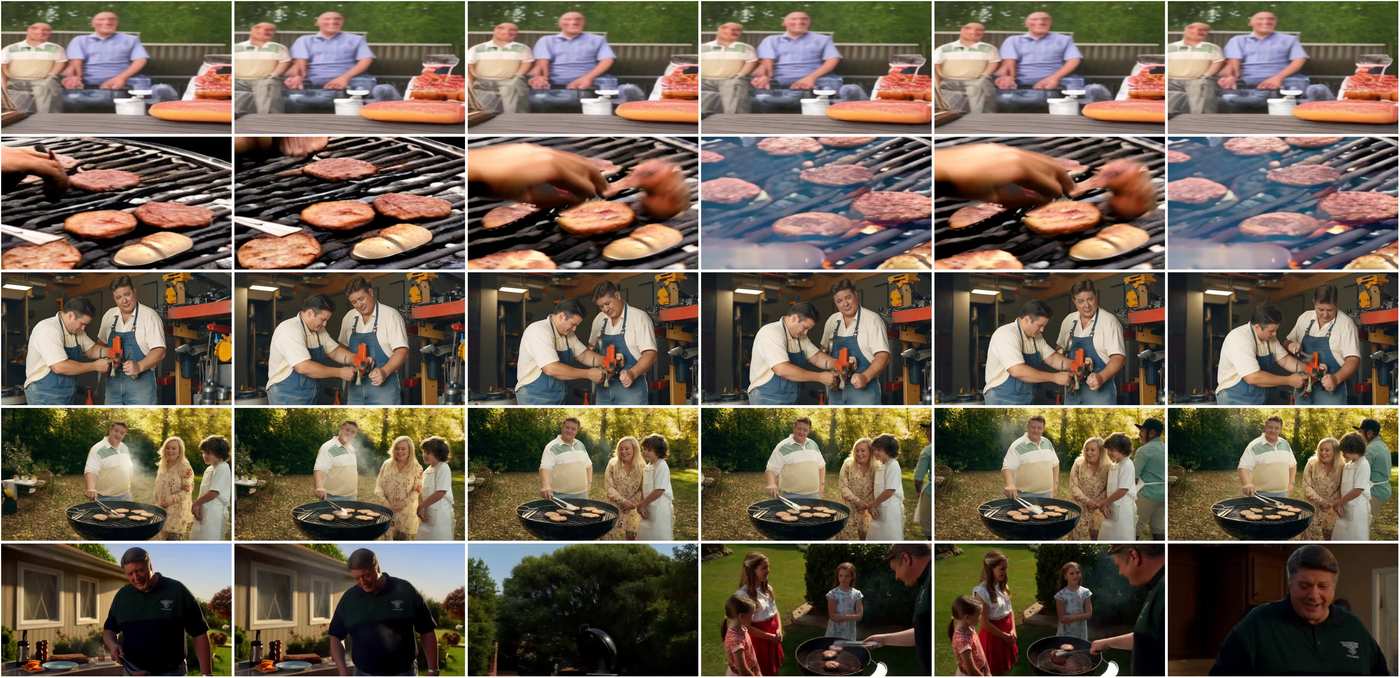}

    \begin{minipage}{0.9\linewidth}
        \scriptsize
        \raggedright
        \textbf{Prompt:} Penny is standing on a stage, nervously holding a microphone, trying to sing while imagining herself as a famous star.
    \end{minipage}
    \includegraphics[width=0.9\linewidth]{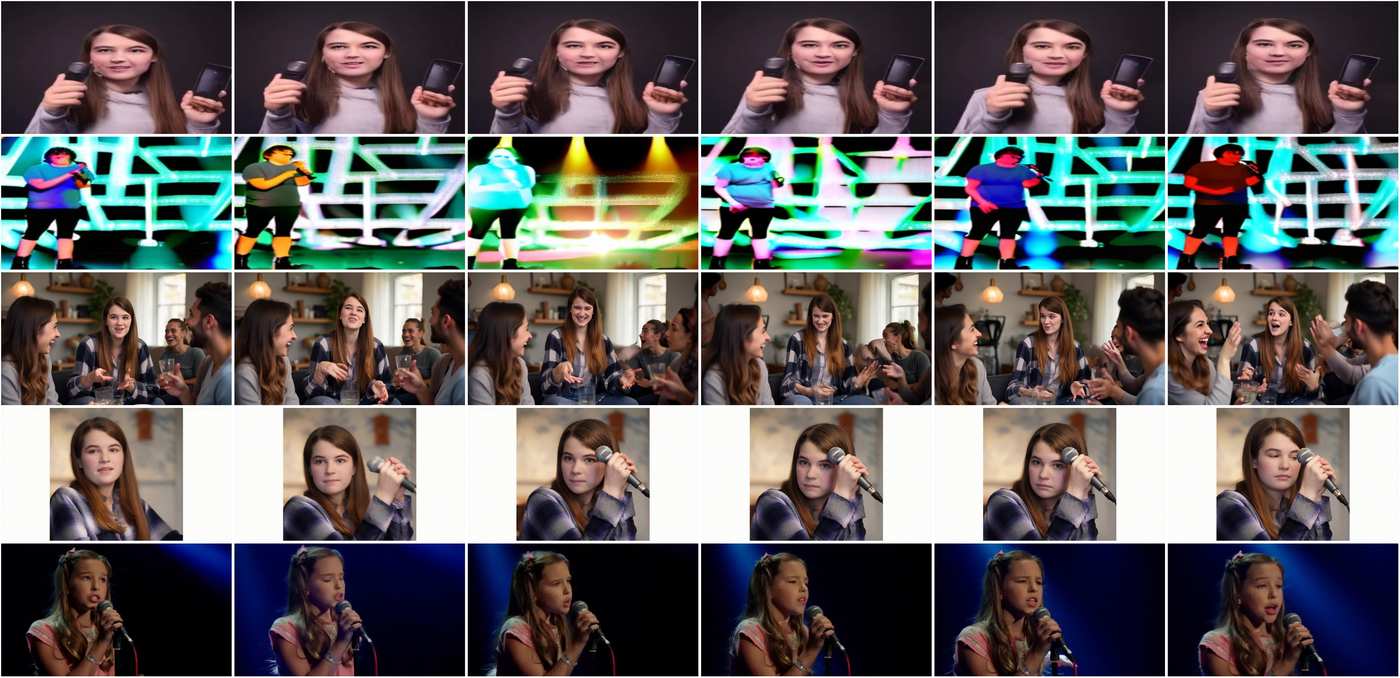}

    \caption{\textbf{Comparison on single-subject generation.} From top to bottom: results from VideoBooth~\cite{videobooth}, DreamVideo, Wan2.1~\cite{wan2025video}, SkyReel-A2~\cite{fei2025skyreelsa2} and ours.}

    \label{fig:supp_compare_sinngle_subject4}
\end{figure}

\end{document}